\documentclass{ecai} 
\usepackage{latexsym}
\usepackage{amssymb}
\usepackage{amsmath}
\usepackage{amsthm}
\usepackage{booktabs}
\usepackage{enumitem}
\usepackage{graphicx}
\usepackage{color}
\usepackage{url}
\usepackage{hyperref}

\usepackage[ruled,vlined]{algorithm2e}
\usepackage[switch]{lineno}
\usepackage{comment}
\usepackage{multirow}
\usepackage{colortbl}
\usepackage{xcolor}
\usepackage{float}
\usepackage{dsfont}
\definecolor{celadon}{rgb}{0.67, 0.88, 0.69}

\usepackage{tabularx}
\usepackage{mathtools}
\usepackage{bm}
\usepackage{bbm}

\usepackage{caption}
\newcommand{\BibTeX}{B\kern-.05em{\sc i\kern-.025em b}\kern-.08em\TeX}

\begin{document}

%%%%%%%%%%%%%%%%%%%%%%%%%%%%%%%%%%%%%%%%%%%%%%%%%%%%%%%%%%%%%%%%%%%%%%%%

\begin{frontmatter}

%%% Use this command to specify your submission number.
%%% In doubleblind mode, it will be printed on the first page.

\paperid{9651} 

%%% Use this command to specify the title of your paper.

\title{Automatic Calibration for Membership Inference Attack on Large Language Models}

%%% Use this combinations of commands to specify all authors of your 
%%% paper. Use \fnms{} and \snm{} to indicate everyone's first names 
%%% and surname. This will help the publisher with indexing the 
%%% proceedings. Please use a reasonable approximation in case your 
%%% name does not neatly split into "first names" and "surname".
%%% Specifying your ORCID digital identifier is optional. 
%%% Use the \thanks{} command to indicate one or more corresponding 
%%% authors and their email address(es). If so desired, you can specify
%%% author contributions using the \footnote{} command.

% \author[A]{\fnms{Saleh}~\snm{Zare Zade}\thanks{Equal contribution.}}
% \author[A]{\fnms{Saleh}~\snm{Zare Zade}\textsuperscript{\dag,}}
\author[A]{\fnms{Saleh}~\snm{Zare Zade}\thanks{Equal Contribution.}}
\author[B,*]{\fnms{Yao}~\snm{Qiang}}
\author[A]{\fnms{Xiangyu}~\snm{Zhou}}
\author[A]{\fnms{Hui}~\snm{Zhu}}
\author[A]{\fnms{Mohammad Amin}~\snm{Roshani}}
\author[A]{\fnms{Prashant}~\snm{Khanduri}}
\author[A]{\fnms{Dongxiao}~\snm{Zhu}}
% \author[A]{\fnms{Dongxiao}~\snm{Zhu}\thanks{Corresponding Author.}}

\address[A]{Department of Computer Science, Wayne State University}
\address[B]{Department of Computer Science, Oakland University}

\address[]{salehz, xiangyu, hui, roshani, khanduri.prashant, dzhu@wayne.edu, qiang@oakland.edu}
%%% Use this environment to include an abstract of your paper.

\begin{abstract}
    Membership Inference Attacks (MIAs) have recently been employed to determine whether a specific text was part of the pre-training data of Large Language Models (LLMs). However, existing methods often misinfer non-members as members, leading to a high false positive rate, or depend on additional reference models for probability calibration, which limits their practicality. To overcome these challenges, we introduce a novel framework called Automatic Calibration Membership Inference Attack (ACMIA), which utilizes a tunable temperature to calibrate output probabilities effectively. This approach is inspired by our theoretical insights into maximum likelihood estimation during the pre-training of LLMs. We introduce ACMIA in three configurations designed to accommodate different levels of model access and increase the probability gap between members and non-members, improving the reliability and robustness of membership inference. Extensive experiments on various open-source LLMs demonstrate that our proposed attack is highly effective, robust, and generalizable, surpassing state-of-the-art baselines across three widely used benchmarks. Our code is available at: \href{https://github.com/Salehzz/ACMIA}{\textcolor{blue}{Github}}.

    % \footnote{Our code is available at: \textcolor{blue}{\url{https://github.com/Salehzz/ACMIA}.}}
\end{abstract}

\end{frontmatter}

%%%%%%%%%%%%%%%%%%%%%%%%%%%%%%%%%%%%%%%%%%%%%%%%%%%%%%%%%%%%%%%%%%%%%%%%

\section{Introduction}
% \textbf{Background} 
% \footnote{\dag\ Equal Contribution.}
% \footnotetext[1]{\dag Equal contribution}
Large Language Models (LLMs), pre-trained on massive text corpora, have shown impressive human-level language understanding, reasoning, and decision-making capabilities \cite{brown2020language,touvron2023llama,achiam2023gpt,roshani2025generative}. However, their tendency to memorize training data also introduces significant ethical and security concerns \cite{liu2024rethinking,wu2024unveiling,biderman2024emergent,qiang2023hijacking,qiang2024learning}. For instance, memorized private information can be vulnerable to privacy breaches  \cite{carlini2021extracting}, while retained copyrighted content, such as news articles, violates the rights of content creators \cite{grynbaum2023times}. Moreover, the likelihood of evaluation data being inadvertently included during training increases, raising concerns about the reliability and validity of evaluation benchmarks \cite{oren2023proving}.
% \begin{figure*} [t]
%   \centering
%   \includegraphics[width=\linewidth]{illustration.png} 
%   \caption{Illustration examples: Let $x_1$ represent a non-training text and $x_2$ a training text. To infer whether a text was included in the training data, Min-$K$\% identifies the $K$\% of tokens with the lowest probabilities, highlighted in red boxes, which led to a false positive for $x_1$. In contrast, our proposed ACMIA method automatically calibrates the probability distribution using a tunable temperature $\tau$, enabling more reliable membership inference and preventing false positives. \label{fig:illustration}}
% \end{figure*}
\begin{figure*}[t]
  \centering
  \includegraphics[width=\linewidth]{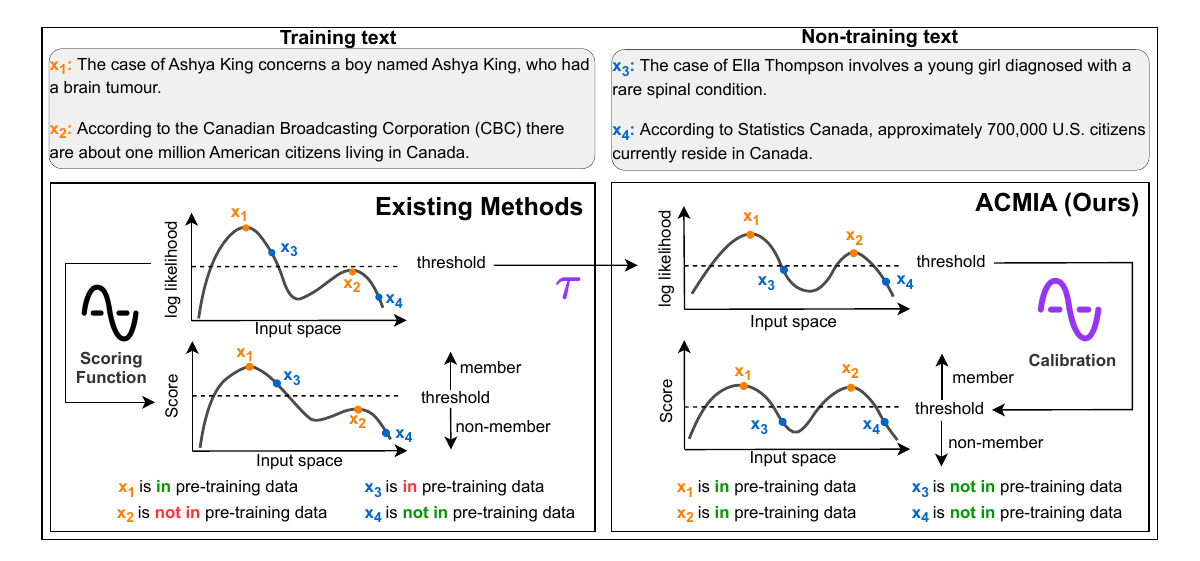} 
  \caption{Illustration of our ACMIA vs. existing methods: Let $x_1$ and $x_2$ represent member (training) texts, and $x_3$ and $x_4$ represent non-member (non-training) texts. Existing methods typically rely on the output log probabilities of input tokens to determine membership, which can result in both false positives (e.g., non-member $x_3$ is classified as member), and false negatives (e.g., member $x_2$ is classified as non-member). In contrast, our proposed ACMIA method first reshapes the log-likelihood distribution using a tunable temperature $\tau$ to enhance the separation between members and non-members. It then calibrates the scores to allow a fair comparison between simple and complex samples, regardless of their membership status. Specifically, among the member texts, $x_1$ represents a simpler example while $x_2$ is more complex; similarly, $x_3$ and $x_4$ are non-member texts, with $x_3$ being simpler and $x_4$ more complex. By adjusting for these differences, our method reduces both false positives and false negatives.
  % It then calibrates the scores to enable fair comparison between simple (high probability such as $x_1$) and complex (low probability such as $x_3$) texts, thereby reducing both false positives and false negatives.
  \label{fig:illustration}}
\end{figure*}

% \textbf{MIAs} 
Membership Inference Attacks (MIAs), originally proposed to determine whether a specific sample was part of the training data \cite{shokri2017membership}, have recently been applied to LLMs to infer whether a given text was used during pre-training. It has significant implications for measuring memorization and privacy risks \cite{carlini2022quantifying,mireshghallah2022quantifying,steinke2024privacy} and detecting evaluation data contamination \cite{oren2023proving} and copyrighted content \cite{duarte2024cop} in LLMs. However, due to the diverse and large-scale nature of training data and the unique training characteristics of LLMs, conducting MIAs against LLMs is significantly more challenging compared to traditional deep learning models \cite{duan2024membership}.

% \textbf{Literature and Reserach Gap} 

Previous MIAs against LLM capitalize on the principle that the content present in pre-training is more likely to be generated compared to the content that is absent \cite{mattern2023membership,shi2023detecting}. Since LLMs tend to assign lower loss values or higher probabilities to training data, typical score-based MIAs exploit these metrics by exploiting the models' tendency to overfit to their training content \cite{shi2023detecting,zhang2024min}. The limitation of these approaches lies in their tendency to misinfer non-members as members, resulting in a high false positive rate, as shown in the example of Figure \ref{fig:illustration}. Specifically, simple and short texts, regardless of whether they originate from members or non-members, are naturally assigned higher probabilities compared to more complex ones \cite{mattern2023membership}. To mitigate this issue, some studies have introduced calibration mechanisms, which refine inference scores accordingly by quantifying the intrinsic complexity of texts. For instance, the final inference scores are derived by averaging results from additional reference models \cite{carlini2022membership}. However, their practicality is limited by the dependency on extra reference models and the additional computational resources required. 

\begin{figure*}[t]
  \centering
  \includegraphics[width=\linewidth]{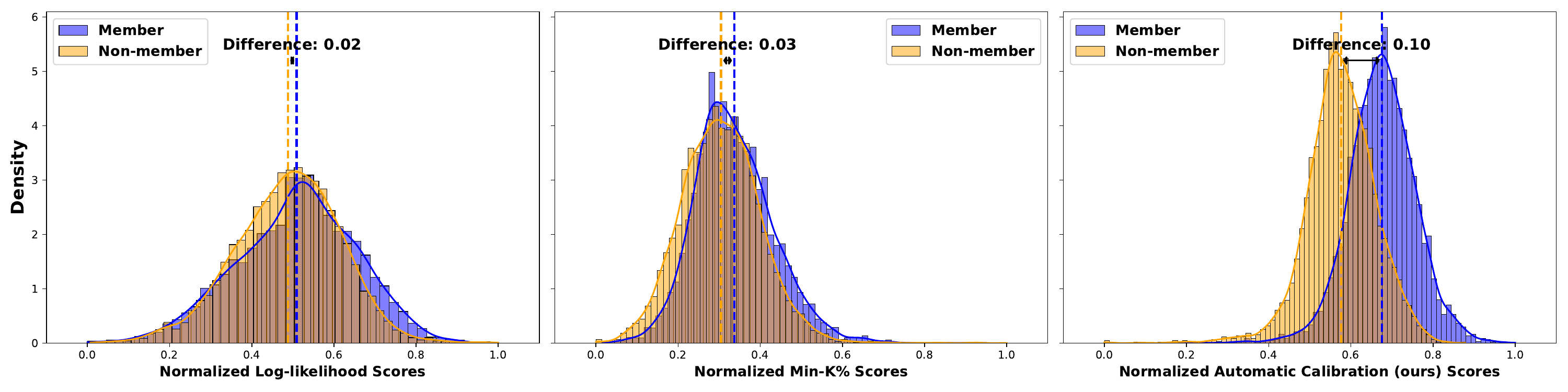} 
  \caption{The three histograms show the density distributions of normalized scores for members (blue) and non-members (orange) across different methods: (left) Log-likelihood scores, (middle) Min-$K$\% scores, and (right) our automatic calibrated scores. The numerical difference between member and non-member means is indicated in each plot, showing that our approach achieves the largest separation, improving the reliability of membership inference. \label{fig:motivation}}
\end{figure*}

Recent works \cite{shi2023detecting,zhang2024min} propose MIA approaches by analyzing the probabilities of the lowest $K\%$ tokens. However, they are constrained by their focus on a limited portion of the text, and as a result they suffer from high false positive rate and false negative rate as illustrated in Figure \ref{fig:illustration}. Consequently, these methods fail to effectively capture the distribution shift between members and non-members during the LLM generation process, as shown in Figure \ref{fig:motivation}. Their performance on datasets where members and non-members share the same distribution is poor, often approaching random guessing.

% \textbf{Our Solution}
To overcome the limitations of existing MIAs on LLMs, we introduce a novel framework called Automatic Calibration Membership Inference Attack (ACMIA), motivated by our theoretical insights into maximum likelihood estimation during the pre-training of LLMs. \textit{Firstly}, to minimize the risk of misinferring non-members as members, we incorporate a temperature-scaled probability with a tunable temperature. Importantly, this temperature adjustment is applied only during post-generation analysis for refining the log likelihood distribution as illustrated in Figure \ref{fig:illustration}. By automatically reshaping the probability distribution of each token in the texts, ACMIA effectively amplifies the probability gap between members and non-members, even when they originate from the same distribution, as illustrated in Figure \ref{fig:motivation}. This enhances the reliability and robustness of membership inference by reducing both false positives and false negatives. \textit{Secondly}, after computing the temperature-scaled probabilities, ACMIA applies calibration techniques to account for the inherent difficulty of each sample, which can otherwise distort membership signals. Traditional reference-based MIAs address this by using an external model to estimate sample difficulty, but ACMIA achieves this by leveraging the target model’s own outputs, without requiring additional models or inference runs. 

We develop three variants of ACMIA, each calibrating the temperature-adjusted probabilities. Two variants rely only on the log probability of the tokens, while the third requires access to the full log probability distribution of each token. This design enables ACMIA to operate effectively under varying levels of model access.

MIAs are widely used as a diagnostic tool to expose potential privacy leakage in pre-trained models. In this context, our proposed method, ACMIA, serves as a red-teaming approach to uncover subtle memorization patterns in LLMs. By providing a more accurate and calibrated assessment of membership status, ACMIA facilitates the identification of training data exposure and supports the design of future privacy-preserving solutions, such as differential privacy and machine unlearning. This promotes greater transparency and accountability in how LLMs handle sensitive or copyrighted content.
% \textbf{Our Findings}

We evaluate the performance of ACMIA with various open-source LLMs, including  Baichuan \cite{Baichuan}, Qwen1.5 \cite{Qwen1.5}, OPT \cite{zhang2022opt}, Pythia \cite{biderman2024emergent}, and GPT-NeoX \cite{black2022gpt} on three benchmarks, i.e., WikiMIA \cite{shi2023detecting}, MIMIR \cite{duan2024membership}, and PatentMIA \cite{zhang2024pretraining}. Our extensive experiments demonstrate that ACMIA achieves state-of-the-art performance across different baselines and models. Additionally, we perform a comprehensive empirical analysis on the effects of temperature, model size, and text length, further demonstrating ACMIA’s effectiveness, robustness, and adaptability across various settings.

\section{Related Work}

% \subsection{Membership Inference Attack on LLMs}
MIAs aim to determine whether a given sample is part of a model’s training data \cite{shokri2017membership,yeom2018privacy}. It has important implications for tasks such as measuring memorization and privacy risks \cite{carlini2022quantifying,mireshghallah2022quantifying,steinke2024privacy}, forming the foundation for advanced attacks \cite{carlini2021extracting,nasr2023scalable}, detecting test set contamination \cite{oren2023proving}, and copyrighted content \cite{meeus2023did,duarte2024cop} in LLMs. Research on MIAs has been investigated in NLP within both fine-tuning \cite{watson2021importance,mireshghallah2022quantifying,fu2023practical,mattern2023membership} and pre-training settings \cite{shi2023detecting,duan2024membership,zhang2024min}. The previous line of work can generally be divided into three categories: score-based methods \cite{shi2023detecting,zhang2024min}, calibration-based methods \cite{watson2021importance,mattern2023membership,zhang2024pretraining}, and reference model methods \cite{carlini2022membership}.

The problem of MIA on LLMs was first explored in \cite{shi2023detecting}, which introduced the WikiMIA benchmark and developed the Min-$K$\% method. Min-$K$\% evaluated the probability of outlier words within a given text, enabling the assessment of whether the text was likely included in the pre-training corpus. Based on Min-$K$\% \cite{zhang2024min} further introduced Min-$K$\%++ by examining whether the input text forms a mode or has a relatively high probability under the conditional categorical distribution. Although these two score-based methods demonstrated improved performance compared to other baselines \cite{yeom2018privacy}, they only focus on a small portion of the text for membership inference and require fine-tuning of the crucial hyper-parameter $K$. This limitation reduces their practicality in real-world applications and their effectiveness on more challenging benchmarks like MIMIR, where there is minimal distribution shift between members and non-members. Several calibration-based methods have been proposed to utilize a difficulty calibration score to regularize raw scores \cite{watson2021importance}. More recently, \cite{zhang2024pretraining} derived the detection score via computing the cross-entropy (i.e., the divergence) between the token probability distribution and the token frequency distribution as a divergence-based calibration method. \cite{carlini2021extracting,ye2022enhanced,mireshghallah2022quantifying} train reference models to rectify anomalies with the average of different models. Other methods, such as \cite{liu2024probing}, utilized the probing technique for pre-training data detection by examining the model’s internal activations. 

Different from previous approaches, this work introduces an automatic calibration method for MIAs on LLMs that does not require any external models. Our approach utilizes a tunable temperature to refine the probability distribution and enhance the distinction between members and non-members. The proposed ACMIA framework achieves state-of-the-art performance across multiple baselines and model architectures, demonstrating strong robustness and generalizability across diverse settings.

\section{Preliminaries}

\subsection{Problem Formulation}

The core task of an MIA is to determine whether a particular sample $x$ was part of the training dataset $\mathcal{D}$ used to train the target LLM $\mathcal{M}$. This task involves calculating an inference score $s(x; \mathcal{M})$, which is further compared against a predefined threshold $\lambda$ to infer whether $x$ is a member of $\mathcal{D}$, i.e., $x \in D$ or not, i.e., $x \notin \mathcal{D}$, formally:
\begin{align}
\label{eq:definition}
    \text{MI}(x,\mathcal{M})=\begin{cases} 
      1, & ~\text{if}~ s(x;\mathcal{M})\ge \lambda \\
      0~, &~ \text{otherwise} 
  \end{cases}.
\end{align}
$\text{MI}(x,\mathcal{M}) = 1$ implies $x\in\mathcal{D}$ and $\text{MI}(x,\mathcal{M}) = 0$ implies $x\notin\mathcal{D}$. The success of MIAs critically depends on the design and precision of the scoring function that best separates training data from non-training data using the threshold $\lambda$. 

\subsection{LLMs}

LLMs are generally trained using maximum likelihood estimation aiming to maximize the probability of the training token sequences \cite{wang2022language}. In this work, we consider $\mathcal{M}$ as an auto-regressive LLM that outputs a probability distribution of the next token given the prefix. Specifically, auto-regressive LLMs apply the chain rule to decompose the probability of a token sequence as: $p(x) = p(x_t|x_1,x_2,...x_{t-1}) \cdot p(x_1,x_2,...x_{t-1})$. For simplicity, we abbreviate the prefix of $x_t$ as $x_{<t}$ throughout the paper. During inference, LLMs generate new tokens sequentially based on the predicted conditional categorical distribution $p(\cdot|x_{<t})$ over the vocabulary.

Following the established standard \cite{shi2023detecting,zhang2024min,duan2024membership}, we consider gray-box access of the target model $\mathcal{M}$, meaning that the adversary can only access the output statistics, including the loss value, logits, and token probabilities. Additional information, such as the model weights and gradients, are not available. 

\section{Method}

\subsection{Motivation}

Based on the Implicit Score Matching (ISM) objective \cite{hyvarinen2005estimation}, $\psi(\bm{x})=\frac{\partial\log p(\bm{x})}{\partial \bm{x}}$ denotes the score function measuring the sensitivity of the log-likelihood with respect to the input text. The maximum likelihood estimation used in the LLM pre-training process can be formulated as
\begin{equation}
    \frac{1}{N}\sum_{\bm{x}}\biggl[\frac{1}{2}||\psi(\bm{x})||^2+\underbrace{\sum_{i=1}^d\frac{\partial\psi_i(\bm{x})}{\partial \bm{x}_i}}_{\mathclap{\substack{\text{the sum of the second-order}\\\text{partial derivatives}}}}\biggr],
    \label{eq:motivation}
\end{equation}
where $\bm{x}_i$ represents the $i$-th dimension in the input $\bm{x}$ of length $d$, and $N$ indicates the number of training samples. For simplicity, the model parameters in the definition of $\psi$ are omitted.
% where $\bm{x}_i$ represents the $i$-th token in a text $\bm{x}$ of length $d$, and $N$ indicates the number of training samples. For simplicity, the model parameters in the definition of $\psi$ are omitted.
Specifically, the maximum likelihood training objective inherently minimizes the magnitude of the first-order derivatives of the likelihood $\log p(\bm{x})$ with respect to $\bm{x}$, as well as the sum of the second-order partial derivatives of $\log p(\bm{x})$ with respect to each dimension of $\bm{x}$.  As a result, the training process encourages smooth likelihood variations around training samples, leading to relatively small gradients in their vicinity. Moreover, since the training procedure maximizes the likelihood of observed data points, it implicitly shapes the likelihood function such that training samples tend to be situated near peaks in the likelihood landscape. Consequently, the first-order derivative $\frac{\partial\log p(\bm{x})}{\partial\bm{x}}$ approaches zero, while the second-order derivative $\frac{\partial^2\log p(\bm{x})}{\partial\bm{x}_i^2}$ is minimized to be negative for the training samples. In other words, the training samples of $M$ are likely to correspond to or be close to the local maximum. Building on this insight, we design an inference mechanism to evaluate whether a sample is a member or a non-member based on its proximity to a local maximum in the output probability space.

\subsection{Automatic Calibration for MIA}

A fundamental challenge in MIA against LLMs is the significant overlap between members and non-members, particularly when sampled from the same distribution. Therefore, both the members and non-members have the chance to lie near a local maximum. In other words, both groups can exhibit similar log-likelihood scores, leading to a high false positive rate where many non-members are misclassified as members. As illustrated in Figure \ref{fig:motivation}, the raw log-likelihood scores and Min-$K$\% scores show minimal separation between members and non-members. To address this challenge, we propose an automatic calibration process that applies a temperature-based transformation, effectively amplifying subtle differences in the underlying distributions. As shown in the rightmost plot of Figure \ref{fig:motivation}, this calibrated approach significantly enhances the separation between members and non-members, reducing the false positive rate and improving the robustness of MIA detection.

Our method dynamically adjusts both first-order and second-order derivatives through temperature scaling, refining the maximum likelihood estimation as:
\begin{equation}
    \frac{1}{N}\sum_{\bm{x}}\biggl[\frac{1}{2\tau^2}||\psi'(\bm{x})||^2+\frac{1}{\tau}\sum_{i=1}^d\frac{\partial\psi_i'(\bm{x})}{\partial \bm{x}_i}\biggr],
    \label{eq:temperature}
\end{equation}
where $\psi'(x) = \tau \cdot \frac{\partial \log p(x, \tau)}{\partial x}$ denotes the scaled gradient of the log probability $\log p(x, \tau)$, with $p(x, \tau)$ representing the temperature-adjusted probability distribution obtained by applying a scaling factor $\tau$ to the logits. Here, the temperature $\tau$ plays a crucial role in balancing the influence of first-order and second-order derivatives with respect to the inputs. When non-members are close to a local maximum, the first-order derivative tends to zero, prompting $\tau$ to shift the emphasis toward the second-order derivative to better capture subtle variations in the likelihood landscape as demonstrated in Figure \ref{fig:illustration}. Conversely, when non-members are far from a local maximum, $\tau$ prioritizes the first-order derivative in the log-likelihood estimation. This dynamic calibration by $\tau$ enhances the robustness of our log-likelihood-based MIA, effectively minimizing false positives and improving inference accuracy. The full derivation of Equation~\ref{eq:temperature} is provided in Section B of the Appendix.

Specifically, We formulate our \textbf{T}emperature-\textbf{S}caled \textbf{P}robabilities (TSP) with temperature $\tau$ as:
\begin{align}
\label{eq:tsp}
\text{TSP}(z\,|\, x_{<t}, \tau) = \frac{\exp\left(\frac{\log p(z\,|\,x_{<t})}{\tau}\right)}{\sum_{i=1}^{K} \exp\left(\frac{\log p(\mathcal{V}_i\,|\,x_{<t})}{\tau}\right)},
\end{align}
where $\tau$ here is the tunable parameter, $\mathcal{V}$ denotes the whole vocabulary with size $K$, and $z$ represents a token in $\mathcal{V}$.

However, merely adjusting the temperature is not enough, as different samples vary in complexity and frequency within the pre-training data of LLMs. In our first version of ACMIA, we construct an internal reference model by applying an adjusted temperature to the target model, such that the resulting log-likelihood serves as the reference, eliminating the reliance on any external models. By tuning the temperature in \eqref{eq:tsp}, we simulate variations in the training process of $\mathcal{M}$. Specifically, setting the temperature $> 1$ simulates a model underfitted on its pre-training data, as it smooths the log-likelihood distribution, leading to increased entropy and lower confidence. Conversely, setting the temperature $< 1$ simulates a model overfitted on its pre-training data, as it sharpens the log-likelihood distribution, leading to reduced entropy and higher confidence. Comparing the target model to this internal reference, we infer that higher probabilities in the overfitted model suggest a sample was likely in the pre-training data. Similarly, higher probabilities in the target model compared to the underfitted model indicate the same. We name this first version as AC and formulate it as: 
\begin{equation}
    \begin{aligned}
    \text{AC}(x,\tau)& = 
    \frac{1}{\lvert \operatorname{FOS}(x) \rvert} \text{sgn}(1-\tau) \\
  & \!\!\!\! \sum_{x_t \in \operatorname{FOS}(x)} \!\! \big(\log \text{TSP}(x_t| x_{<t}, \tau) -\log p(x_t | x_{<t})\big),
    \end{aligned}
\end{equation}
where $\text{sgn}(1-\tau)$ determines whether the temperature calibration sharpens or smooths the log-likelihood distribution. $\operatorname{FOS}(x)$ represents the first occurrence of tokens in a sentence, limiting calculations to these tokens since later repetitions are easier for the LLM to predict \cite{zhang2024pretraining}.

Our second version of ACMIA named DerivAC is proposed to include the derivative of the temperature-scaled log probability with respect to temperature. This derivative provides a fairer score based on the complexity of a sample to determine its membership. Specifically, a higher derivative indicates that the sample is more likely to belong to the pre-training data. The intuition is that as the temperature increases, if the probability of the sample also increases, it suggests the sample is near a local maximum. We then formulate DerivAC as:
\begin{equation}
    \begin{aligned}
    \text{DerivAC}(x,\tau) = 
    \frac{1}{\lvert \operatorname{FOS}(x) \rvert}& \sum_{x_t \in \operatorname{FOS}(x)} \\
    (\log \text{TSP}(x_t| x_{<t}, \tau + \delta)
                    & -\log \text{TSP}(x_t| x_{<t}, \tau) ),
    \end{aligned}
\end{equation}
% \begin{equation}
%     \begin{aligned}
%     \text{DerivAC}(x,\tau) = 
%     \frac{1}{\lvert \operatorname{FOS}(x) \rvert}& \sum_{x_t \in \operatorname{FOS}(x)}
%     \log \frac{\text{TLP}(x_t| x_{<t}, \tau + \delta)}{\text{TLP}(x_t| x_{<t}, \tau)}
%     \end{aligned}
% \end{equation}
where $\delta > 0$ is a constant that remains the same for all scores, eliminating the need to divide the equation by $\delta$. The derivative can be computed directly without involving $\delta$, as detailed in the Section B of Appendix. 

\begin{table*}[t]
\caption{AUROC scores for evaluating MIAs across different benchmarks and LLMs. Higher scores indicate better detection of pre-training texts. We report the AUROC scores of our ACMIA methods, i.e., AC, DerivAC, and NormAC, with the optimal temperature. We report the average AUROC scores of different subsets in MIMIR. \textbf{Bold} indicates the best MIA performance.}
\centering
\small % font size
\setlength{\tabcolsep}{7.pt} % column separation
\renewcommand{\arraystretch}{1.1} % row space 
\begin{tabular}{l cccc cccccc cccc}
\toprule
\multirow{4}{*}{\textbf{Method}}
& \multicolumn{4}{c}{PatentMIA}
& \multicolumn{6}{c}{WikiMIA}
& \multicolumn{4}{c}{MIMIR}
\\ 
\cmidrule(r){2-5}
\cmidrule(r){6-11}
\cmidrule(r){12-15}
% \cmidrule{5-8} 
% & Baichuan-13B   & Qwen1.5-14B   & OPT-6.7B & Pythia-6.9B & Pythia-12B & OPT-13B & GPT-NeoX-20B
% & \multirow{2}{*}{\textbf{B-13B}}
% & \multirow{2}{*}{\textbf{B2-13B}}
% & \multirow{2}{*}{\textbf{Q-32B}}
% & \multirow{2}{*}{\textbf{Q-72B}}
& \multicolumn{2}{c}{\textbf{Baichuan}}
& \multicolumn{2}{c}{\textbf{Qwen1.5}}

& \multicolumn{2}{c}{\textbf{OPT-6.7B}}
% & \multicolumn{2}{c}{\textbf{Pythia-6.9B}}
& \multicolumn{2}{c}{\textbf{Pythia-12B}}
% & \multicolumn{2}{c}{\textbf{OPT-13B}}
& \multicolumn{2}{c}{\textbf{NeoX-20B}}

& \multicolumn{4}{c}{\textbf{Pythia}}
\\ 
\cmidrule(r){2-3}
\cmidrule(r){4-5}
\cmidrule(r){6-7}
\cmidrule(r){8-9}
\cmidrule(r){10-11}
\cmidrule(r){12-15}
% \cmidrule(r){12-13}
% \cmidrule(r){14-15}

& 13B & 2-13B & 32B & 72B & \textit{Ori.} & \textit{Para.} & \textit{Ori.} & \textit{Para.} & \textit{Ori.} & \textit{Para.} & 1.4B & 2.8B & 6.9B & 12B
\\ 
\midrule
Loss
& 60.8 & 63.0 & 60.9 & 54.7
& 62.4    & 61.5    & 65.3    & 65.1    & 70.4    & 69.3
& 80.2 & 80.5 & 81.2 & 81.6

\\
Ref
& 60.4 & 47.4 & 57.6 & 34.2
& 63.6    & 63.7    & 62.2    & 60.6    & 68.0    & 67.7
& 65.9 & 63.9 & 64.9 & 64.1

\\
Lowercase
& - & - & - & -
& 58.4    & 57.4    & 60.4    & 60.0    & 66.8    & 66.7
& 76.3 & 78.1 & 79.0 & 78.8

\\
Zlib
& 63.6 & 67.6 & 63.6 & 53.5
& 64.3    & 64.3    & 67.5    & 67.7    & 72.0    & 71.8
& 77.4 & 77.9 & 78.6 & 78.9

\\
Min-K\%
& 66.7 & 70.1 & 67.5 & 59.1
& 67.4    & 64.7    & 69.8    & 67.8    & 75.5    & 72.3
& 80.6 & 80.9 & 81.7 & 82.2

\\
Min-K\%++
& 62.8 & 66.6 & 66.6 & 65.1
& 69.0    & 64.9    & 71.4    & 67.8    & 75.4    & 71.8
& 71.4 & 73.2 & 74.0 & 75.5

\\
DC-PDD
& 70.0 & 74.7 & 70.1 & 61.4
& 67.9    & 66.1    & 70.1    & 68.1    & 75.8    & 73.1
& 81.4 & 81.7 & 82.3 & 82.5

\\
\hline
AC
& 73.4 & 78.0 & 77.0 & 80.1
& 70.1    & 68.0    & 72.6    & 70.3    & \textbf{78.2}    & 75.8
& 81.7 & \textbf{82.3} & 83.0 & \textbf{83.6}

\\
DerivAC
& 75.1 & 78.3 & \textbf{77.9} & \textbf{81.4}
& 70.2    & 68.1    & 72.6    & 70.3    & \textbf{78.2}    & \textbf{75.9}
& \textbf{81.8} & \textbf{82.3} & \textbf{83.2} & 83.5

\\
NormAC
& \textbf{76.5} & \textbf{78.5} & 77.7 & 80.8
& \textbf{71.4}    & \textbf{70.6}    & \textbf{74.1}    & \textbf{71.3}    & 78.1    & 74.9
& \textbf{81.8} & 82.1 & 83.1 & 83.2

\\
\bottomrule
\end{tabular}

\label{tab:auroc}
\end{table*}

Furthermore, to develop a calibration method that treats samples fairly based on their difficulty, normalization using mean and variance offers an effective solution. Instead of relying on a specific distribution (e.g., temperature = 1), this method identifies the distribution that best distinguishes members from non-members after normalization. The objective is to position member samples closer to local maxima while pushing non-member samples farther away. This method named NormAC is formulated as:
% \begin{equation}
%     \begin{aligned}
%     \text{NormAC}(x,\tau) = &\frac{1}{\lvert \operatorname{FOS}(x) \rvert} \sum_{x_t \in \operatorname{FOS}(x)} \\
%     & \frac{\log \text{TLP}(x_t| x_{<t}, \tau) -\mu_{x_t}} {\sigma_{x_t}},
%     \end{aligned}
% \end{equation}
\begin{equation}
    \begin{aligned}
    \text{NormAC}(x,\tau) = \frac{1}{\lvert \operatorname{FOS}(x) \rvert} \sum_{x_t \in \operatorname{FOS}(x)} \frac{\log \text{TSP}(x_t| x_{<t}, \tau) -\mu_{x_t}} {\sigma_{x_t}},
    \end{aligned}
\end{equation}
where 
\begin{align}
    \mu_{x_t} = \mathbb{E}_{z \sim \text{TSP}(\cdot | x_{<t},\tau)}[\log \text{TSP}(z| x_{<t},\tau)],
\end{align}
and 
\begin{align}
    \sigma_{x_t} = \sqrt{\mathbb{E}_{z \sim \text{TSP}(\cdot | x_{<t}, \tau)}[(\log \text{TSP}(z| x_{<t},\tau)-\mu_{x_t})^{2}]}.
\end{align}
Finally, the scores generated by ACMIA methods, including AC, DerivAC, and NormAC, are used to infer whether the target sample $x$ is originated from the training data. It is noteworthy that the first two versions of ACMIA (AC and DerivAC) do not require access to the full log-likelihood distribution; they only need the capability to fine-tune the temperature $\tau$, which is usually feasible even with API access. Conversely, the third version (NormAC) relies on access to the token log probability distribution to compute the mean and variance, making it dependent on a higher level of model access. Furthermore, We discuss in the Section C of Appendix how AC and DerivAC can also be implemented using only the loss (negative log-likelihood) of the sample, making them applicable in even more restricted settings.

\begin{table*}[t]
\caption{TPR@5\%FPR scores for evaluating MIAs across different benchmarks and LLMs. Higher scores indicate better detection of pre-training texts. We report the TPR@5\%FPR scores of our ACMIA methods, i.e., AC, DerivAC, and NormAC, with the optimal temperature. We report the average TPR@5\%FPR scores of different subsets in MIMIR. More detailed results of these subsets are reported in the Appendix. \textbf{Bold} indicates the best MIA performance.}
\centering
\small % font size
\setlength{\tabcolsep}{7.pt} % column separation
\renewcommand{\arraystretch}{1.1} % row space 
\begin{tabular}{l cccc cccccc cccc}
\toprule
\multirow{4}{*}{\textbf{Method}}
& \multicolumn{4}{c}{PatentMIA}
& \multicolumn{6}{c}{WikiMIA}
& \multicolumn{4}{c}{MIMIR}
\\ 
\cmidrule(r){2-5}
\cmidrule(r){6-11}
\cmidrule(r){12-15}
% \cmidrule{5-8} 
% & Baichuan-13B   & Qwen1.5-14B   & OPT-6.7B & Pythia-6.9B & Pythia-12B & OPT-13B & GPT-NeoX-20B
% & \multirow{2}{*}{\textbf{B-13B}}
% & \multirow{2}{*}{\textbf{B2-13B}}
% & \multirow{2}{*}{\textbf{Q-32B}}
% & \multirow{2}{*}{\textbf{Q-72B}}
& \multicolumn{2}{c}{\textbf{Baichuan}}
& \multicolumn{2}{c}{\textbf{Qwen1.5}}

& \multicolumn{2}{c}{\textbf{OPT-6.7B}}
% & \multicolumn{2}{c}{\textbf{Pythia-6.9B}}
& \multicolumn{2}{c}{\textbf{Pythia-12B}}
% & \multicolumn{2}{c}{\textbf{OPT-13B}}
& \multicolumn{2}{c}{\textbf{NeoX-20B}}

& \multicolumn{4}{c}{\textbf{Pythia}}
\\ 
\cmidrule(r){2-3}
\cmidrule(r){4-5}
\cmidrule(r){6-7}
\cmidrule(r){8-9}
\cmidrule(r){10-11}
\cmidrule(r){12-15}
% \cmidrule(r){12-13}
% \cmidrule(r){14-15}

& 13B & 2-13B & 32B & 72B & \textit{Ori.} & \textit{Para.} & \textit{Ori.} & \textit{Para.} & \textit{Ori.} & \textit{Para.} & 1.4B & 2.8B & 6.9B & 12B
\\ 
\midrule
Loss
& 16.3 & 18.6 & 17.0 & 10.8
& 12.9    & 15.1    & 15.8    & 18.7    & 20.1    & 18.0
& 41.3 & 41.1 & 43.0 & 43.7

\\
Ref
& 14.0 & 3.4 & 7.7 & 1.4
& 10.8    & 8.6    & 9.4    & 10.1    & 15.8    & 18.7
& 15.3 & 14.7 & 15.9 & 14.4

\\
Lowercase
& - & - & - & -
& 10.1    & 7.9    & 13.7    & 11.5    & 12.9    & 12.9
& 41.2 & 39.8 & 40.8 & 40.2

\\
Zlib
& 16.1 & 16.8 & 13.4 & 6.3
& 18.0    & 15.1    & 23.7    & 19.4    & 21.6    & 22.3
& 33.4 & 34.3 & 34.2 & 35.2

\\
Min-K\%
& 17.3 & 21.7 & 18.3 & 10.9
& 20.9    & 15.8    & 23.7    & 20.9    & 24.5    & 22.3
& 43.9 & 44.8 & 45.8 & 47.3

\\
Min-K\%++
& 13.6 & 17.0 & 16.1 & 14.8
& 23.7    & 21.6    & 23.0    & 22.3    & 24.5    & 18.7
& 28.5 & 30.0 & 31.2 & 31.3

\\
DC-PDD
& \textbf{28.0} & \textbf{27.4} & 22.3 & 11.0
& 19.4    & 21.6    & 25.9    & 21.6    & \textbf{30.2}    & \textbf{28.1}
& 50.5 & 51.2 & 51.2 & 51.8

\\
\hline
AC
& 23.8 & 25.8 & 29.7 & 33.0
& \textbf{33.1}    & 30.9    & 27.3    & \textbf{26.6}    & \textbf{30.2}    & 27.3
& 51.6 & 51.8 & 53.2 & 52.9

\\
DerivAC
& 25.9 & 25.2 & \textbf{30.6} & \textbf{33.6}
& \textbf{33.1}    & 30.9    & 27.3    & \textbf{26.6}    & 28.8    & 27.3
& 53.6 & \textbf{53.9} & 54.3 & \textbf{54.6}

\\
NormAC
& 26.3 & 24.3 & 24.7 & 25.7
& 30.2    & \textbf{35.3}    & \textbf{28.8}    & \textbf{26.6}    & 28.1    & 25.2
& \textbf{54.1} & 53.8 & \textbf{55.4} & 54.3

\\
\bottomrule
\end{tabular}

\label{tab:tpr}
\end{table*}
\begin{table}[h]
\caption{FPR@95\%TPR scores for evaluating MIAs on PatentMIA across different LLMs. Lower scores indicate better detection of pre-training texts. \textbf{Bold} indicates the best MIA performance.}
\centering
\small % font size
\setlength{\tabcolsep}{4.pt} % column separation
\renewcommand{\arraystretch}{1.1} % row space 
\resizebox{0.48\textwidth}{!}{%
\begin{tabular}{l cc ccccccc}
\toprule
\multirow{2}{*}{\textbf{Method}}
& \multicolumn{2}{c}{\textbf{Baichuan}}
& \multicolumn{7}{c}{\textbf{Qwen1.5}}
\\ 
\cmidrule(r){2-3}
\cmidrule(r){4-10}
% \cmidrule(r){12-13}
% \cmidrule(r){14-15}

& 13B & 2-13B & 0.5B & 1.8B & 4B & 7B & 14B & 32B & 72B
\\ 
\midrule
Loss
& 94.6 & 94.4 & 95.5 & 95.6 & 95.5 & 95.3 & 95.2 & 94.9 & 95.7

\\
Zlib
& 90.0 & 84.9 & 94.2 & 93.6 & 92.4 & 91.1 & 89.5 & 88.0 & 91.2

\\
Min-K\%
& 83.8 & 81.7 & 90.6 & 89.7 & 87.6 & 86.8 & 84.0 & 83.3 & 88.1

\\
Min-K\%++
& 89.8 & 84.5 & 91.3 & 90.0 & 89.7 & 89.7 & 88.7 & 85.4 & 88.1

\\
DC-PDD
& 93.8 & 85.6 & 93.0 & 97.1 & 95.2 & 94.3 & 93.5 & 91.0 & 93.4

\\
\hline
AC
& 83.1 & 71.5 & 88.7 & 89.0 & 86.3 & 84.1 & 83.2 & 76.6 & 69.7

\\
DerivAC
& 80.8 & 67.0 & 83.7 & 84.8 & 81.2 & 79.1 & 78.9 & 72.9 & 66.1

\\
NormAC
& \textbf{72.9} & \textbf{61.8} & \textbf{76.2} & \textbf{77.6} & \textbf{73.9} & \textbf{72.8} & \textbf{71.9} & \textbf{66.2} & \textbf{61.2}

\\
\bottomrule
\end{tabular}
\label{tab:patent_fpr}
}

\end{table}

\section{Experiment Settings}

\subsection{Datasets}

Our experiments involve three main benchmarks for evaluating MIAs:  WikiMIA \cite{shi2023detecting}, MIMIR \cite{duan2024membership}, and PatentMIA \cite{zhang2024pretraining}. \textbf{WikiMIA} organizes Wikipedia event texts by publication date into training and non-training data, further split by sentence length for fine-grained evaluation. It includes original (verbatim texts) and paraphrased (ChatGPT-generated) settings to evaluate the robustness of detection performance. \textbf{MIMIR}, sourced from the Pile dataset \cite{gao2020pile}, draws training and non-training samples from identical distributions within their respective sets. This setup is more challenging than WikiMIA due to minimal distribution shifts and temporal discrepancies \cite{duan2024membership}. \textbf{PatentMIA} is a Chinese-language benchmark for evaluating pre-training data detection beyond English. To differentiate training from non-training data, the benchmark leverages the priority dates of patents, for instance, texts with priority dates after the release of the LLM are guaranteed to be excluded from its pre-training.
\vspace{-0.05in}
\subsection{LLMs}

We evaluate our method and baselines on open-source LLMs, including Pythia (12B) \cite{biderman2024emergent}, OPT (6.7B) \cite{zhang2022opt}, and GPT-NeoX (20B) \cite{black2022gpt} on WikiMIA. For MIMIR, we focus on the Pythia family, analyzing variants from 160M to 12B parameters for consistency with \cite{duan2024membership}. We evaluate MIA approaches against Baichuan-13B \cite{Baichuan}, Baichuan2-13B, and the Qwen1.5 family \cite{Qwen1.5}, representing key models for Chinese text generation, on PatentMIA. We apply the smallest available models and the best-performing reference models for each type, such as OPT-350M for OPT and Pythia-70M for Pythia, ensuring consistency with previous studies.
\vspace{-0.05in}
\subsection{Baselines}

In evaluating our ACMIA on LLMs, we compare its performance with several advanced baselines. Specifically, we examine three different score-based methods: Loss \cite{yeom2018privacy}, Min-$K$\% \cite{shi2023detecting}, and Min-$K$\%++ \cite{zhang2024min}. Additionally, Lowercase and Zlib \cite{carlini2021extracting} are proposed enhancements that leverage calibration to refine the original scoring functions. Furthermore, Ref \cite{carlini2021extracting} aims to improve the precision of the Loss attack and reduce the false negative rate by incorporating an external model, which is distinct from the target model. DC-PDD \cite{zhang2024pretraining} is a divergence-based calibration method using cross-entropy between token probabilities and token frequencies for improving detection scores. A detailed description of all baselines is provided in Section A of the Appendix.

\subsection{Metrics}

Following \cite{shi2023detecting,zhang2024min}, our experiments utilize the area under the receiver operating characteristic curve (AUROC) as the major evaluation metric. This metric indicates the likelihood that the model accurately differentiates between members (``positive") and non-members (``negative"). Specifically, the AUROC offers a threshold-independent evaluation of inference capability, where a score above the 50\% random-guessing baseline signifies an improved performance. Additionally, we evaluate the true positive rate at 5\% false positive rate (TPR@5\%FPR) and the false positive rate at 95\% true positive rate (FPR@95\%TPR) to measure the MIAs' accuracy under strict thresholds, ensuring the relevance and rigor of our evaluations.

\subsection{Implementation Details.}

The temperature $\tau$ is tuned through an exponential function, formally: $\tau = e^{\alpha}$, allowing us to tune the temperature by optimizing $\alpha$. This exponential formulation allows for a continuous and smooth adjustment of the temperature, enabling precise control over the model's output log probability distribution. Furthermore, we can explore a wide range of temperature values systematically, from very low to very high, within a compact numerical range by using $e^{\alpha}$. This tuning method further improves the efficiency of hyper-parameter tuning by quickly converging to the optimal temperature. For different datasets, we find the optimal temperature using a set of labeled member and non-member samples, consistent with the assumptions made by existing baseline methods. For example, Min-K\% and Min-K\%++ involve tuning the hyperparameter K, Ref method rely on choosing a suitable external model, and DC-PDD involves tuning an upper bound hyperparameter, selecting a reference corpus, and accessing the model’s vocabulary to estimate token frequencies.

\section{Results and Analysis}

\subsection{ACMIA Result}

The AUROC results presented in Table \ref{tab:auroc} provide a comprehensive evaluation of different MIA methods on three benchmarks with different LLMs. Notably, different versions of ACMIA, i.e., AC, DerivAC, and NormAC, consistently outperform baselines, demonstrating their effectiveness in distinguishing training data from non-training data across all benchmarks and models. 

The baselines, including Loss, Min-$K$\%, and Min-$K$\%++, show significantly weaker performance, particularly in PatentMIA and WikiMIA, where AUROC scores remain below 70. These results indicate that these score-based methods struggle to effectively infer training data from non-training data in these challenging benchmarks. Additionally, reference model approaches, such as Ref, consistently under-perform, indicating that the reliance on static reference models limits their effectiveness in adapting to LLM-induced distribution shifts. In contrast, our proposed ACMIAs, which eliminate the reliance on reference models and incorporate adaptive temperature calibration, prove to be more efficient while attaining much better performance. The latest advanced baseline, DC-PDD, outperforms other baselines. However, it depends on an external reference corpus, which may not generalize effectively across diverse data structures, as evidenced by the high variance in performance across different benchmarks. While all the ACMIA variants consistently deliver superior performance across various benchmarks without requiring any external resources.

\begin{figure}[t]
  \centering
  \includegraphics[width=\linewidth]{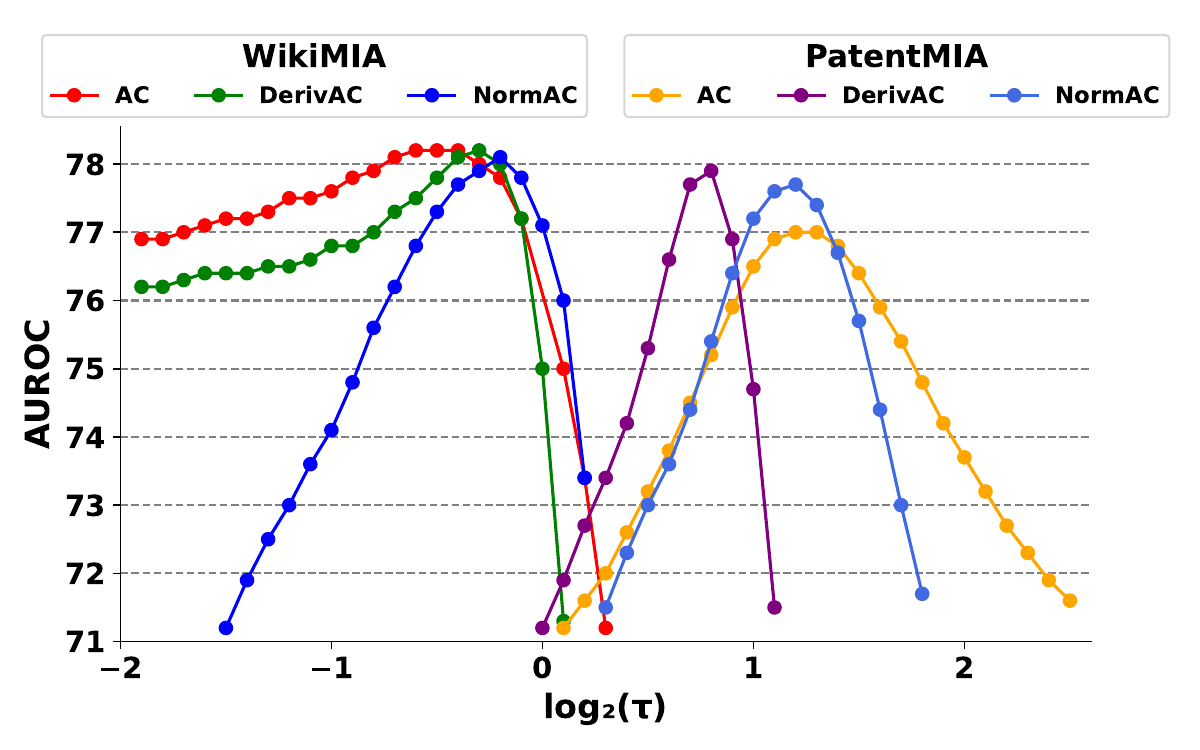} 
  \caption{AUROC with different temperatures on WikiMIA using NeoX-20B and PatentMIA using Qwen1.5-32B. Our ACMIA framework, including AC, DerivAC, and NormAC, is fairly robust to the tunable temperatures.\label{fig:temp}}
\end{figure}

The TPR@5\%FPR results in Table \ref{tab:tpr} provide further insight into the performance of different MIAs. Our proposed approaches once again outperform all baselines, demonstrating higher detection accuracy while controlling for false positives. Most baselines show significantly lower TPR@5\%FPR scores, often below 20 across most settings. This indicates their limited effectiveness in correctly identifying training data while maintaining a strict false positive rate of 5\%. DC-PDD stands out as the strongest baseline method, outperforming other traditional MIAs. It achieves notably high scores in PatentMIA (28.0 on Baichuan-13B, 27.4 on Baichuan2-13B) and NeoX-20B (30.2, WikiMIA original). These results indicate that DC-PDD’s divergence-based calibration technique is effective in maintaining a strong balance between precision and recall, making it a viable alternative to basic score-based attacks. However, it fails to outperform our ACMIA methods, particularly on WikiMIA and MIMIR benchmarks, where temperature-based calibration proves to be a more flexible and effective adjustment mechanism. These results clearly demonstrate the effectiveness of ACMIA in reducing the false positive rate, particularly on the most challenging dataset, MIMIR, where it achieves significant success.

Furthermore, the consistent performance across original and paraphrased text settings in WikiMIA, as shown in Table \ref{tab:auroc} and Table \ref{tab:tpr}, highlights the robustness of our proposed ACMIA methods to text modifications. The minimal performance drop between original and paraphrased settings suggests that ACMIA methods effectively capture deeper structural and probabilistic signals that persist despite lexical variations, making them significantly more adaptable than traditional MIA approaches.

Additionally, the FPR@95\%TPR results in Table~\ref{tab:patent_fpr} show that most baseline methods perform close to random guessing, with false positive rates near 95\% when evaluated on complex training samples. This indicates their limited ability to distinguish such samples from non-members under strict conditions where the true positive rate must remain at or above 95\%. In contrast, our proposed methods, particularly NormAC, achieve significantly lower false positive rates, demonstrating improved robustness and effectiveness in identifying challenging member instances.

A clear model size performance correlation is evident across all evaluated baseline methods in Table \ref{tab:auroc}, Table \ref{tab:tpr}, and Table \ref{tab:patent_fpr}. Larger models, such as Pythia-12B and NeoX-20B, consistently achieve higher AUROC and TPR@5\%FPR scores, reinforcing the trend that larger LLMs exhibit greater levels of memorization and are, therefore, more susceptible to MIAs. As LLMs continue to scale in size and complexity, this trend highlights the urgent need for more effective privacy-preserving techniques to mitigate the risks associated with training data leakage.

\subsection{Ablation Study}

\textbf{Ablation on the temperature:} Figure \ref{fig:temp} presents the AUROC performance of different versions of ACMIA, i.e., AC, DerivAC, and NormAC, on WikiMIA and PatentMIA under varying temperature settings. The results indicate that our ACMIA approaches exhibit strong robustness to temperature variations, maintaining high AUROC scores over a broad range of temperature values. This suggests that our methods do not rely on precise temperature tuning and can generalize well across different conditions, making them highly practical and adaptable.

\noindent\textbf{Ablation on the optimal temperature across models:} Table \ref{tab:patent_temp} presents the optimal temperature values ($\log_2\tau$) for different ACMIA versions on the PatentMIA dataset. The results show that these values are consistent across models of varying sizes trained on similar pre-training data, indicating that the optimal temperature is determined more by the dataset than by the model.

\begin{table}[t]
\caption{Optimal $\log_2\tau$ for AC, DerivAC, and NormAC on PatentMIA dataset.}
\centering
\small % font size
\setlength{\tabcolsep}{3.pt} % column separation
\renewcommand{\arraystretch}{1.1} % row space 
\resizebox{0.48\textwidth}{!}{%
\begin{tabular}{l cc ccccccc}
\toprule
\multirow{2}{*}{\textbf{Method}}
& \multicolumn{2}{c}{\textbf{Baichuan}}
& \multicolumn{7}{c}{\textbf{Qwen1.5}}
\\ 
\cmidrule(r){2-3}
\cmidrule(r){4-10}
% \cmidrule(r){12-13}
% \cmidrule(r){14-15}

& 13B & 2-13B & 0.5B & 1.8B & 4B & 7B & 14B & 32B & 72B
\\ 
\midrule

$\log_2\tau$ AC
& 1.4 & 1.3 & 1.4 & 1.3 & 1.2 & 1.2 & 1.2 & 1.3 & 1.1

\\
$\log_2\tau$ DerivAC
& 0.9 & 0.7 & 0.8 & 0.8 & 0.7 & 0.7 & 0.7 & 0.8 & 0.8

\\
$\log_2\tau$ NormAC
& 1.3 & 1.1 & 1.4 & 1.2 & 1.1 & 1.1 & 1.1 & 1.2 & 1.1

\\
\bottomrule
\end{tabular}
\label{tab:patent_temp}
}

\end{table}

\begin{figure}[t]
  \centering
  \includegraphics[width=\linewidth]{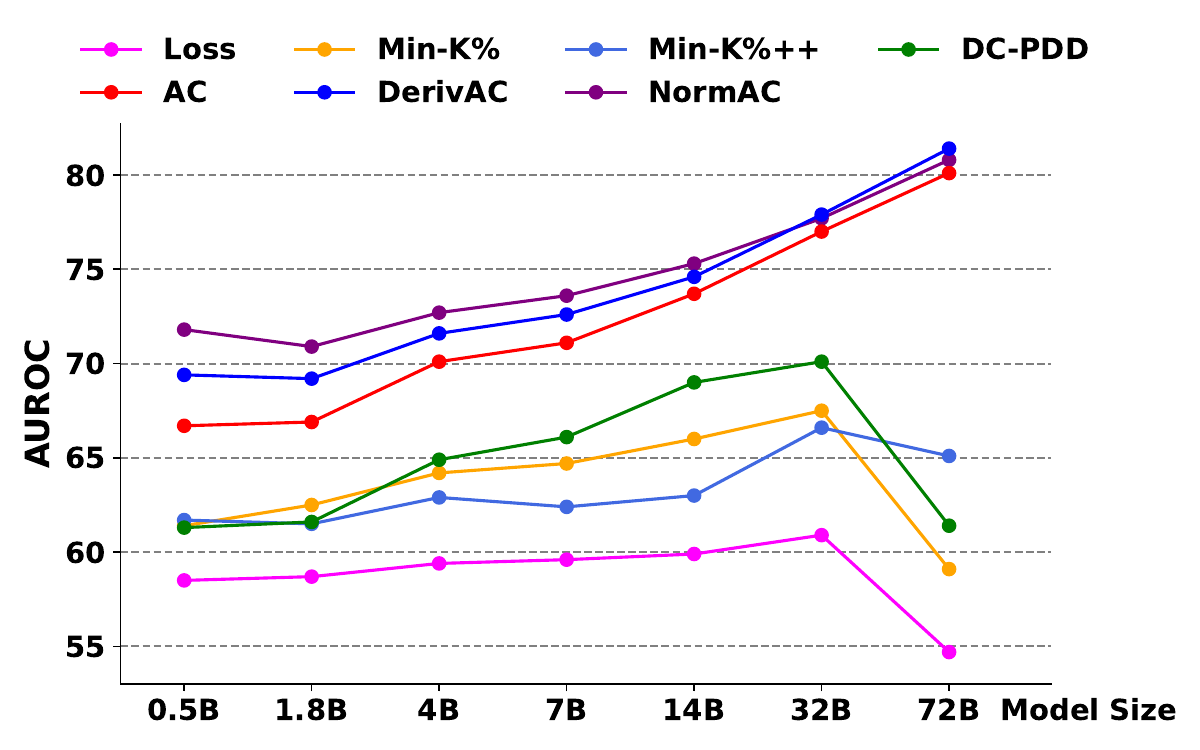} 
  \caption{AUROC with different sizes of Qwen1.5 on PatentMIA.}
  \vspace{+0.1in}
  \label{fig:model_size}
\end{figure}

\noindent
\textbf{Ablation on the model size:} Figure \ref{fig:model_size} demonstrates the relationship between model size and AUROC performance for various MIA methods on PatentMIA using different sizes of Qwen1.5. The results reveal a clear trend where larger models exhibit higher AUROC scores, indicating that larger models are more vulnerable to MIAs. This aligns with prior findings that larger LLMs tend to memorize training data more effectively, making them easier to distinguish from non-training data.  Furthermore, our methods, i.e.,AC, DerivAC, and NormAC, consistently outperform other baselines across different model sizes, proving to be the most effective MIAs.

\begin{figure}[t]
  \centering
  \includegraphics[width=\linewidth]{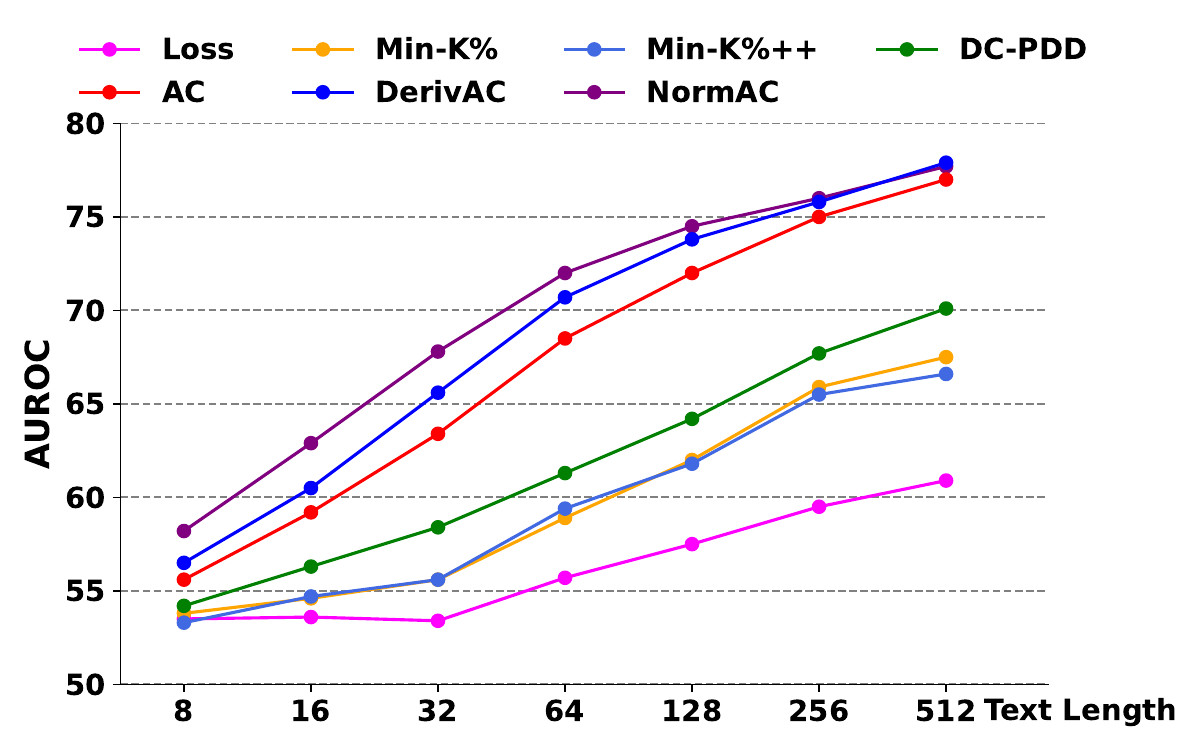}
  \caption{AUROC with different text lengths on PatentMIA dataset.}
  \vspace{+0.1in}
  \label{fig:text_size}
\end{figure}

\noindent
\textbf{Ablation on the text size:} Figure \ref{fig:text_size} illustrates how AUROC performance evolves with increasing text length across various MIA methods on the PatentMIA benchmark. It is clear that longer text lengths significantly improve MIA effectiveness, with all methods achieving higher AUROC scores as the text size increases. Consistently, our proposed ACMIAs outperform all other baselines under different text length settings. These results highlight the growing vulnerability of LLMs to membership inference as input length increases, emphasizing the need for stronger privacy protections for longer sequences, e,g., the whole copyrighted news report.

%\section{Limitations}
%Our method, like most existing MIA approaches such as Min-K\%, Min-K\%++, and DC-PDD, assumes access to token-level probabilities or log probabilities. This assumption may not hold in fully closed-source LLMs, such as ChatGPT. However, our focus is on open-source, white-box, or gray-box settings, where such access is typically available through APIs or model outputs. 

%\section{Future Direction}
\section{Conclusion}

This work presents a novel and efficient automatic calibration approach to enhance MIA. Our ACMIA effectively enhances the distinction between members and non-members, thereby reducing false positive and false negative rates. Our extensive experiments on multiple open-source LLMs and benchmarks demonstrate that ACMIA consistently outperforms the existing methods, achieving state-of-the-art results while maintaining adaptability across different settings. Our method is grounded in the principles of maximum likelihood estimation, making it potentially applicable to a broader range of generative and classification models trained under the similar objective. Extending ACMIA to these settings could be a promising direction for future research.
% Even in black-box setting where token-level probabilities are unavailable, it can be adapted using the overall negative log-likelihood of the input sequence, as long as the model permits temperature adjustment during inference. This flexibility broadens the applicability of ACMIA while addressing common access constraints in baseline methods.

%Our method, based on maximum likelihood estimation, is applicable to a wide range of generative and classification models trained under the same objective, with future research extending ACMIA to these settings. 

% \section{Ethics Statement}
% The main goal of our work is to offer a framework for understanding vulnerabilities in LLMs against membership inference attacks, rather than creating new opportunities for malicious activity. Our work has significant societal impact by raising public awareness of LLM security and safety while promoting the adoption of new defenses to address these risks.
%%%%%%%%%%%%%%%%%%%%

%%% Use this command to include your bibliography file.
\clearpage
\bibliography{Arxiv}

%%% Appendix.

\clearpage
\appendix

\section{Baselines Details}
\label{sec:baselines}
We consider seven MIAs from the three categories as the baselines. Below are further details:

\noindent
\textbf{Score-based methods:} 

(1) \textbf{Loss} \cite{yeom2018privacy} considers the model’s computed loss over the target sample as the membership inference score: 
% $s(x; \mathcal{M}) = \mathcal{L}(x; \mathcal{M})$.
\begin{equation}
    s(x; \mathcal{M}) = \mathcal{L}(x; \mathcal{M}).
\end{equation}

(2) \textbf{Min-K\%} \cite{shi2023detecting} calculates a score using the $K$\% of tokens with the lowest likelihood rather than averaging all token probabilities as:
% $s(x; \mathcal{M}) = \frac{1}{|\text{min}-k(x)|}\sum_{x_i \in \text{min}-k(x)} -\log(p(x_i|x_1,\cdots,x_{i-1}))$. 
\begin{equation}
    s(x; \mathcal{M}) = \frac{1}{|\text{Min-}k\%(x)|}\sum_{x_i \in \text{Min-}k\%(x)} -\log(p(x_i|x_1,\cdots,x_{i-1})).
\end{equation}

(3) \textbf{Min-K\%++} \cite{zhang2024min} examines whether the training data $x$ has a relatively high probability under the conditional categorical distribution $p(\cdot|x_{<t})$, formally: 
\begin{align}
\text{Min-K\%++}(x_{<t},x_t)&=\frac{\log p(x_t|x_{<t})-\mu_{x_{<t}}}{\sigma_{x_{<t}}},
\end{align}

\begin{align}
\label{eq:mink++}
s(x; \mathcal{M})&=\frac{1}{|\text{min-k\%}|}\sum_{x_t\in\text{min-k\%}}\text{Min-K\%++}(x_{<t},x_t).
\end{align}
Here, $\mu_{x_{<t}}=\mathbb{E}_{z\sim p(\cdot|x_{<t})}[\log p(z|x_{<t})]$ is the expectation of the next token's log probability over the vocabulary given the prefix $x_{<t}$, and $\sigma_{x_{<t}}=\text{sqrt}(\mathbb{E}_{z\sim p(\cdot|x_{<t})}[(\log p(z|x_{<t})-\mu_{x_{<t}})^2])$ is the standard deviation.

\noindent
\textbf{Calibration-based methods:}

(4) \textbf{Zlib Entropy} \cite{carlini2021extracting} calibrates the sample’s loss under the target model using the sample’s zlib compression size, formally: 
\begin{equation}
    s(x; \mathcal{M}) = \frac{\mathcal{L}(x; \mathcal{M})}{zlib(x)}.
\end{equation}

(5) \textbf{Lowercase} \cite{carlini2021extracting} detects memorization by comparing a model’s perplexity on the original input versus a canonicalized version of it. Specifically, it measures the ratio between the perplexity of the input and that of its lowercased variant, under the same model:
\begin{equation}
    s(x; \mathcal{M}) = \frac{\mathcal{L}(\text{Lowercase}(x); \mathcal{M})}{\mathcal{L}(x; \mathcal{M})}.
\end{equation}
This ratio highlights memorization patterns where the model is sensitive to the original casing, which is often preserved in training data.

%(1) \textbf{Neighborhood attack} \cite{mattern2023membership} uses an estimate of the curvature of the loss function at a given sample, which is computed by perturbing the target sequence to create $n$ `neighboring’ points and comparing the loss of the target $x$ with its neighbors $\Tilde{x}$:  
%\begin{equation}
%    s(x; \mathcal{M}) = \mathcal{L}(x; \mathcal{M}) - \frac{1}{n}\sum_{i=1}^n \mathcal{L}(\Tilde{x}_i; \mathcal{M}).
%\end{equation}

\noindent
\textbf{Reference-based methods:}

(6) \textbf{Ref} \cite{carlini2021extracting} attempts to improve on the Loss attack’s precision and reduce the false negative rate by accounting for the intrinsic complexity of the target point $x$ by calibrating $\mathcal{L}(x; \mathcal{M})$, with respect to another reference model $\mathcal{M}_\text{ref}$, which is trained on data from the same distribution as $\mathcal{D}$, but not necessarily the same data: 
\begin{equation}
   s(x; \mathcal{M}) = \mathcal{L}(x; \mathcal{M}) - \mathcal{L}(x; \mathcal{M}_\text{ref}).
\end{equation}

(7) \textbf{DC-PDD}~\cite{zhang2024pretraining} calibrates token probabilities by comparing them against a token frequency distribution derived from a reference corpus. The intuition is that non-training texts with many common tokens may yield misleadingly high likelihoods under the target LLM. DC-PDD computes a divergence-based score that reflects how informative the tokens are relative to their frequency in the reference corpus. The final score is based only on the first occurrence of each token and clipped by a tunable upper bound $a$ to avoid domination by outliers:
\begin{equation}
\begin{aligned}
s(x; \mathcal{M}, D') = & \frac{1}{|\operatorname{FOS}(x)|}\\
& \sum_{x_i \in \operatorname{FOS}(x)} \min\left(-p(x_i; \mathcal{M}) \cdot \log p(x_i;D'), a\right),
\end{aligned}
\end{equation}
where $p(x_i; \mathcal{M})$ is the model-predicted token probability and $p(x_i;D')$ is its empirical frequency in a public reference corpus $D'$.

\section{Proofs}

\subsection{Derivation of Score Function}

The log probability for input $x$ after applying temperature scaling is:
\[
\log p(x, \tau) = \log \frac{\exp(z / \tau)}{\sum_j \exp(z_j / \tau)},
\]
where:
\begin{itemize}
    \item $z$ is the logit corresponding to the input $x$,
    \item $z_j$ are the logits for other vocabulary tokens,
    \item $\tau > 0$ is the temperature.
\end{itemize}
The score function $\psi(x, \tau)$ is the gradient of the log probability with respect to the input $x$:
\[
\psi(x, \tau) = \frac{\partial \log p(x, \tau)}{\partial x}.
\]
Substituting $\log p(x, \tau)$:
\[
\psi(x, \tau) = \frac{\partial}{\partial x} \left( \frac{z}{\tau} - \log \sum_j \exp(z_j / \tau) \right).
\]
The derivative  of $\frac{z}{\tau}$:
\[
\frac{\partial}{\partial x} \frac{z}{\tau} = \frac{1}{\tau} \frac{\partial z}{\partial x}.
\]
The derivative  of $-\log \sum_j \exp(z_j / \tau)$:
\[
\frac{\partial}{\partial x} \left( -\log \sum_j \exp(z_j / \tau) \right) = -\frac{1}{\tau} \sum_j p_j(x, \tau) \frac{\partial z_j}{\partial x}.
\]
The score function is:
\[
\psi(x, \tau) = \frac{1}{\tau} \frac{\partial z}{\partial x} - \frac{1}{\tau} \sum_j p_j(x, \tau) \frac{\partial z_j}{\partial x}.
\]
Here, we define $\psi'(x) = \tau\times\psi(x, \tau)$, so we reach the following:
\[
\psi'(x) = \boxed{\frac{\partial z}{\partial x}} - \sum_j p_j(x, \tau) \frac{\partial z_j}{\partial x}.
\]
Now, we compute the second derivative of the $i$-th component of $\psi(x, \tau)$, $\psi_i(x, \tau)$, with respect to $x_i$.
\[
\frac{\partial}{\partial x_i} \left( \frac{1}{\tau} \frac{\partial z}{\partial x_i} \right) = \frac{1}{\tau} \frac{\partial^2 z}{\partial x_i^2}.
\]
\begin{align*}
\frac{\partial}{\partial x_i} \left( -\frac{1}{\tau} \sum_j p_j(x, \tau) \frac{\partial z_j}{\partial x_i} \right) 
&= -\frac{1}{\tau} \sum_j \Bigg( \frac{\partial p_j(x, \tau)}{\partial x_i} \frac{\partial z_j}{\partial x_i} \\
&\quad + p_j(x, \tau) \frac{\partial^2 z_j}{\partial x_i^2} \Bigg).
\end{align*}
where,
\[
\frac{\partial p_j(x, \tau)}{\partial x_i} = p_j(x, \tau) \left( \frac{1}{\tau} \frac{\partial z_j}{\partial x_i} - \frac{1}{\tau} \sum_k p_k(x, \tau) \frac{\partial z_k}{\partial x_i} \right).
\]
The second derivative of the score function is:
\begin{align*}
\frac{\partial \psi_i(x, \tau)}{\partial x_i} &= \frac{1}{\tau} \frac{\partial^2 z}{\partial x_i^2} 
- \frac{1}{\tau^2} \sum_j p_j(x, \tau) \Bigg( \left( \frac{\partial z_j}{\partial x_i} \right)^2 \\
&\quad - \sum_k p_k(x, \tau) \frac{\partial z_k}{\partial x_i} \frac{\partial z_j}{\partial x_i} \Bigg) \\
&\quad - \frac{1}{\tau} \sum_j p_j(x, \tau) \frac{\partial^2 z_j}{\partial x_i^2}.
\end{align*}
By substituting $\psi'(x) = \tau\times\psi(x, \tau)$, we obtain the following equation:
\begin{align*}
\frac{\partial \psi'_i(x)}{\partial x_i} &= \boxed{\frac{\partial^2 z}{\partial x_i^2}} 
- \frac{1}{\tau} \sum_j p_j(x, \tau) \Bigg( \left( \frac{\partial z_j}{\partial x_i} \right)^2 \\
&\quad - \sum_k p_k(x, \tau) \frac{\partial z_k}{\partial x_i} \frac{\partial z_j}{\partial x_i} \Bigg) \\
&\quad - \sum_j p_j(x, \tau) \frac{\partial^2 z_j}{\partial x_i^2}.
\end{align*}
So after applying temperature, we obtain the following the score function:
\[
\boxed{\frac{1}{N}\sum_{\bm{x}}\biggl[\frac{1}{2\tau^2}||\psi'(\bm{x})||^2+\frac{1}{\tau}\sum_{i=1}^d\frac{\partial\psi_i'(\bm{x})}{\partial \bm{x}_i}\biggr].}
\]

\subsection{Derivation of Log Probability with Temperature Scaling W.R.T Temperature}

The probability \( p(x, \tau) \) is given by the softmax function:
\[
p(x, \tau) = \frac{\exp\left(\frac{z}{\tau}\right)}{\sum_{j} \exp\left(\frac{z_j}{\tau}\right)},
\]
Taking the logarithm of \( p(x, \tau) \), we have:
\[
\log p(x, \tau) = \frac{z}{\tau} - \log\left(\sum_{j} \exp\left(\frac{z_j}{\tau}\right)\right).
\]
To compute the derivative of \( \log p(x, \tau) \) with respect to \( \tau \), we have:
The first term is:
\[
\frac{\partial}{\partial \tau} \left(\frac{z}{\tau}\right) = -\frac{z}{\tau^2}.
\]

\[
\frac{\partial}{\partial \tau} \log\left(\sum_{j} \exp\left(\frac{z_j}{\tau}\right)\right) = \frac{-\frac{1}{\tau^2} \sum_{j} z_j \exp\left(\frac{z_j}{\tau}\right)}{\sum_{j} \exp\left(\frac{z_j}{\tau}\right)}.
\]
Let \( p_j(x, \tau) = \frac{\exp\left(\frac{z_j}{\tau}\right)}{\sum_{k} \exp\left(\frac{z_k}{\tau}\right)} \), which is the probability for token \( j \) after scaling logits using temperature. Then:
\[
\frac{\partial}{\partial \tau} \log\left(\sum_{j} \exp\left(\frac{z_j}{\tau}\right)\right) = -\frac{\mu_z}{\tau^2},
\]
where
\[
\mu_z = \sum_{j} p_j(x, \tau) z_j,
\]
is the softmax-weighted mean of logits.
\[
\frac{\partial}{\partial \tau} \log p(x, \tau) = \frac{\mu_z - z}{\tau^2}.
\]
The first derivative of \( \log p(x, \tau) \) with respect to \( \tau \) is:
\[
\boxed{\frac{\partial}{\partial \tau} -\log p(x, \tau) = \frac{z - \mu_z}{\tau^2}},
\]
where \( \mu_z = \sum_{j} p_j(x, \tau) z_j \) is the probability-weighted mean of the logits.

The first derivative with respect to the temperature \( \tau \) essentially represents the difference between the logit \( z \) for the input token and the softmax-weighted mean of all logits (\( \mu_z \)), normalized by \( \tau^2 \). This indicates that the derivative captures how much the specific logit \( z \) deviates from the average behavior of all logits under the current temperature scaling.

\section{Applying ACMIA Under Limited Model Access}
In this section, we explain how the AC and DerivAC variants of ACMIA can be applied even when only the loss (negative log-likelihood) of each sample is available, without requiring access to token-level log-probabilities. This setting reflects a stricter access level, which is often encountered when working with commercial APIs. Notably, baseline methods such as Min-K\%, Min-K\%++, and DC-PDD are not applicable in this scenario: Min-K\% requires access to the log-probability of each token, Min-K\%++ needs the full log-probability distribution for each token, and DC-PDD relies on both tokens log-probabilities and access to the model’s vocabulary.

We originally defined the AC score as:
\begin{align*}
\text{AC}(x,\tau)& = 
\frac{1}{\lvert \operatorname{FOS}(x) \rvert} \text{sgn}(1-\tau) \\
& \!\!\!\! \sum_{x_t \in \operatorname{FOS}(x)} \!\! \big(\log \text{TSP}(x_t| x_{<t}, \tau) -\log p(x_t | x_{<t})\big),
\end{align*}
where $\operatorname{FOS}(x)$ denotes the set of first occurrences of each token in the input $x$. Now that we do not have access to each token log probability, we need to remove this from our formulation and do our formulation on all of the tokens instead.

To support strict-access scenarios, we replace the token-level formulation with a sample-level version:
\begin{align*}
& \text{AC}(x,\tau) = \text{sgn}(1-\tau)\\
& \!\!\!\! \big(\frac{1}{|x|}\sum_{x_t \in x}\log \text{TSP}(x_t| x_{<t}, \tau)\big) -\big(\frac{1}{|x|}\sum_{x_t \in x}\log p(x_t | x_{<t})\big),
\end{align*}
where
\[
\frac{1}{|x|}\sum_{x_t \in x}\log \text{TSP}(x_t| x_{<t}, \tau)
\]
and 
\[
\frac{1}{|x|}\sum_{x_t \in x}\log p(x_t | x_{<t})
\]
represent the average log-likelihood of the sample $x$ under the temperature-adjusted target model and target model, respectively. This can be compactly rewritten as:
\[
\text{AC}(x,\tau; \mathcal{M}) = \text{sgn}(1-\tau)\big(\mathcal{L}(x; \mathcal{M}) - \mathcal{L}(x; \mathcal{M}_\tau)\big)
\]
where $\mathcal{L}(x; \mathcal{M})$ denotes the target model’s loss (i.e., negative log-likelihood) on sample $x$ using the default temperature $\tau = 1$, and $\mathcal{L}(x; \mathcal{M}_\tau)$ represents the loss computed by the same model with a modified temperature $\tau$, denoted as $\mathcal{M}_\tau$.

A similar approach can be applied to DerivAC by computing a finite-difference approximation of the derivative with respect to temperature using the sample-level losses from models with slightly perturbed temperature values.

We originally defined the DerivAC as:
\begin{align*}
\text{DerivAC}(x,\tau) = 
\frac{1}{\lvert \operatorname{FOS}(x) \rvert}& \sum_{x_t \in \operatorname{FOS}(x)} \\
(\log \text{TSP}(x_t| x_{<t}, \tau + \delta)
                & -\log \text{TSP}(x_t| x_{<t}, \tau) ),
\end{align*}

We replace the token-level formulation with the following sample-level equivalent:
\begin{align*}
\text{DerivAC}(x,\tau) = 
& \big(\frac{1}{|x|}\sum_{x_t \in x}\log \text{TSP}(x_t| x_{<t}, \tau + \delta)\big)\\
& -\big(\frac{1}{|x|}\sum_{x_t \in x}\log \text{TSP}(x_t| x_{<t}, \tau)\big),
\end{align*}
where
\[
\frac{1}{|x|}\sum_{x_t \in x}\log \text{TSP}(x_t| x_{<t}, \tau + \delta)
\]
and 
\[
\frac{1}{|x|}\sum_{x_t \in x}\log \text{TSP}(x_t| x_{<t}, \tau)
\]
represent the average log-likelihood of the sample $x$ under the temperature-adjusted models with temperatures $\tau + \delta$ and $\tau$, respectively. This can be compactly rewritten as:
\[
\text{DerivAC}(x, \tau; \mathcal{M}) = \mathcal{L}(x; \mathcal{M}_{\tau + \delta}) - \mathcal{L}(x; \mathcal{M}_\tau)
\]
where $\mathcal{L}(x; \mathcal{M}_{\tau + \delta})$ denotes the target model’s loss (i.e., negative log-likelihood) on sample $x$ with temperature $\tau + \delta$, and $\mathcal{L}(x; \mathcal{M}_\tau)$ is the corresponding loss computed using the temperature-scaled model $\mathcal{M}_\tau$.

\section{Ethics Statement}
The main goal of our work is to offer a framework for understanding vulnerabilities in LLMs against membership inference attacks, rather than creating new opportunities for malicious activity. Our work has significant societal impact by raising public awareness of LLM security and safety while promoting the adoption of new defenses to address these risks.

\section{Additional Results}
Here we provide additional results.
MIMIR benchmark includes three different settings, each defined by a threshold on the percentage of $n$-gram overlap between member and non-member samples. A higher overlap makes the membership inference task more difficult, as non-members become harder to distinguish from members. For example, a $\leq 20\%$ 7-gram overlap threshold means that each non-member sample shares at most 20\% of its 7-grams with any member sample in the training set. Formally, the $n$-gram overlap for a non-member sample $x = x_1x_2\ldots x_m$ is defined as:
\[
\frac{1}{m - n + 1} \sum_{i=1}^{m - n + 1} 1 \left\{\exists y \in D : x_i\ldots x_{i+n-1} \in y \right\},
\]
where $D$ is the set of member samples and the indicator checks whether each $n$-gram in $x$ appears in any member. This metric captures lexical similarity and allows controlled evaluation of MIA difficulty under varying levels of overlap.

Table~\ref{tab:mimir_auroc_7_20} presents the AUROC results, and Table~\ref{tab:mimir_tpr_7_20} shows the TPR@5\%FPR, both evaluated on the MIMIR benchmark where the overlap between non-member and member samples is limited to less than or equal to 20\% of 7-grams. Our methods outperform the baselines in most subsets and achieve the highest average AUROC and TPR@5\%FPR, indicating strong effectiveness in separating member and non-member samples in low-overlap scenarios.

Table~\ref{tab:mimir_auroc_13_80} presents the AUROC results, and Table~\ref{tab:mimir_tpr_13_80} shows the TPR@5\%FPR, both evaluated on the MIMIR benchmark where non-member samples have less than 80\% 13-gram overlap with member samples. In this setting, our methods consistently outperform the baselines on most subsets. Notably, in Table~\ref{tab:mimir_tpr_13_80}, Our methods achieve, on average, approximately 10\% higher TPR@5\%FPR compared to the best baseline, highlighting their robustness. These results show that ACMIA remains effective at distinguishing member from non-member samples even under high-overlap conditions.

Table~\ref{tab:mimir_auroc_13_20} presents the AUROC results, and Table~\ref{tab:mimir_tpr_13_20} shows the TPR@5\%FPR scores, both evaluated on the MIMIR benchmark where non-member samples have less than 20\% 13-gram overlap with member samples. Once again, our methods achieve the best performance, demonstrating their effectiveness across diverse overlap scenarios.
\begin{table*}[b]
\caption{AUROC results on the challenging MIMIR benchmark \protect\cite{duan2024membership} with non-member sets at $\le 20\%$ 7-gram overlap threshold with a suite of Pythia models \protect\cite{biderman2024emergent}. In each column, the best result across all methods is \textbf{bolded}. Despite not requiring a reference model like the Ref method does, our methods (AC, DerivAC, and NormAC) consistently achieve superior or comparable performance.}
\label{tab:mimir_auroc_7_20}
\begin{center} \scriptsize
\setlength{\tabcolsep}{0.7pt}
\begin{tabularx}{\textwidth}{l *{20}{>{\centering\arraybackslash}X}@{}}
\toprule
\multirow{2}{*}{}  & \multicolumn{5}{c}{\textbf{Wikipedia}} & \multicolumn{5}{c}{\textbf{Github}} & \multicolumn{5}{c}{\textbf{PubMed Central}} \\
\cmidrule(lr){2-6}  \cmidrule(lr){7-11} \cmidrule(lr){12-16}
\textbf{Method}
& 160M & 1.4B & 2.8B & 6.9B & 12B
& 160M & 1.4B & 2.8B & 6.9B & 12B
& 160M & 1.4B & 2.8B & 6.9B & 12B
% & 160M & 1.4B & 2.8B & 6.9B & 12B
\\
\midrule
Loss \cite{yeom2018privacy}
& 62.6 & 65.6 & 66.3 & 67.8 & 68.8 & 84.4 & 87.3 & 88.0 & 88.9 & 89.4 & 78.9 & 78.5 & 78.1 & 78.2 & 78.2
\\
Ref \cite{carlini2021extracting}
& 55.2 & 65.0 & \textbf{68.0} & \textbf{71.9} & \textbf{73.8} & 76.3 & 78.2 & 73.8 & 72.7 & 69.1 & 69.2 & 67.7 & 62.9 & 62.8 & 60.5
\\
Lowercase \cite{carlini2021extracting}
& 61.8 & 64.4 & 64.8 & 66.9 & 67.3 & 83.4 & 86.0 & 86.9 & 88.1 & 89.1 & 72.0 & 73.2 & 72.4 & 72.8 & 72.3
\\
Zlib \cite{carlini2021extracting}
& 56.5 & 61.8 & 63.0 & 64.8 & 65.8 & 87.8 & 90.0 & 90.7 & 91.3 & 91.7 & 77.3 & 77.4 & 77.1 & 77.3 & 77.3
\\
Min-K\% \cite{shi2023detecting}
& 62.6 & 65.8 & 66.6 & 68.2 & 69.5 & 84.4 & 87.3 & 88.1 & 88.9 & 89.5 & 79.3 & 79.4 & 78.9 & 79.2 & 79.4
\\
Min-K\%++ \cite{zhang2024min}
& 60.2 & 64.4 & 65.6 & 68.4 & 70.5 & 79.6 & 83.6 & 85.6 & 85.9 & 87.4 & 65.8 & 68.3 & 68.7 & 69.8 & 70.2
\\
DC-PDD \cite{zhang2024pretraining}
& \textbf{63.3} & 66.1 & 66.9 & 68.6 & 69.4 & \textbf{88.9} & \textbf{91.1} & \textbf{91.7} & \textbf{92.2} & \textbf{92.5} & 81.8 & 80.4 & 80.2 & 80.4 & 79.9
\\
\hline
AC
& 62.9 & \textbf{66.4} & 67.1 & 69.6 & 71.4 & 88.5 & 91.0 & 91.6 & \textbf{92.2} & \textbf{92.5} & 82.3 & 81.0 & 80.8 & 80.7 & 80.7
\\
DerivAC
& 62.9 & \textbf{66.4} & 67.1 & 69.5 & 71.4 & \textbf{88.9} & 90.7 & 91.3 & 92.1 & \textbf{92.5} & \textbf{83.2} & 82.3 & 82.2 & 82.0 & 81.7
\\
NormAC
& 62.2 & 64.9 & 65.6 & 68.9 & 70.3 & 88.2 & 90.4 & 91.1 & 91.5 & 91.8 & 82.0 & \textbf{82.8} & \textbf{82.4} & \textbf{82.7} & \textbf{82.0}
\\
\vspace{-.6em} \\
\toprule
\multirow{2}{*}{}  & \multicolumn{5}{c}{\textbf{ArXiv}} & \multicolumn{5}{c}{\textbf{DM Mathematics}} & \multicolumn{5}{c}{\textbf{Average}}\\
\cmidrule(lr){2-6}  \cmidrule(lr){7-11} \cmidrule(lr){12-16}
\textbf{Method} 
& 160M & 1.4B & 2.8B & 6.9B & 12B
& 160M & 1.4B & 2.8B & 6.9B & 12B
& 160M & 1.4B & 2.8B & 6.9B & 12B
% & 160M & 1.4B & 2.8B & 6.9B & 12B
\\
\midrule
Loss \cite{yeom2018privacy} 
& 75.1 & 77.6 & 78.0 & 79.0 & 79.5 & 94.2 & 92.2 & 91.9 & 92.0 & 91.9 & 79.0 & 80.2 & 80.5 & 81.2 & 81.6
\\
Ref \cite{carlini2021extracting} 
& 63.7 & 70.8 & 71.2 & 73.6 & 74.3 & 76.2 & 47.7 & 43.7 & 43.3 & 42.6 & 68.1 & 65.9 & 63.9 & 64.9 & 64.1
\\
Lowercase \cite{carlini2021extracting}
& 70.3 & 72.5 & 73.7 & 74.6 & 74.8 & 86.1 & 85.3 & 92.8 & 92.8 & 90.3 & 74.7 & 76.3 & 78.1 & 79.0 & 78.8
\\
Zlib \cite{carlini2021extracting}
& 74.2 & 77.1 & 77.5 & 78.3 & 78.6 & 80.4 & 80.8 & 81.2 & 81.2 & 81.1 & 75.2 & 77.4 & 77.9 & 78.6 & 78.9
\\
Min-K\% \cite{shi2023detecting}
& 75.1 & 77.6 & 78.0 & 79.0 & 79.5 & \textbf{94.3} & \textbf{92.9} & 93.1 & 93.0 & 93.1 & 79.1 & 80.6 & 80.9 & 81.7 & 82.2
\\
Min-K\%++ \cite{zhang2024min}
& 62.2 & 65.6 & 67.7 & 69.5 & 71.4 & 81.3 & 75.3 & 78.3 & 76.5 & 77.9 & 69.8 & 71.4 & 73.2 & 74.0 & 75.5
\\
DC-PDD \cite{zhang2024pretraining}
& 75.6 & 78.2 & 78.3 & 79.1 & 79.5 & 91.4 & 91.2 & 91.4 & 91.1 & 91.4 & 80.2 & 81.4 & 81.7 & 82.3 & 82.5
\\
\hline
AC
& 73.9 & 77.3 & 77.9 & 79.2 & 80.0 & 91.0 & 92.8 & \textbf{94.0} & \textbf{93.4} & \textbf{93.4} & 79.7 & 81.7 & \textbf{82.3} & 83.0 & \textbf{83.6}
\\
DerivAC
& 77.1 & 79.2 & 79.3 & 80.3 & 80.8 & 89.5 & 90.6 & 91.4 & 92.0 & 91.1 & \textbf{80.3} & \textbf{81.8} & \textbf{82.3} & \textbf{83.2} & 83.5
\\
NormAC
& \textbf{78.2} & \textbf{80.2} & \textbf{80.5} & \textbf{81.1} & \textbf{81.0} & 88.6 & 90.6 & 91.1 & 91.4 & 90.9 & 79.8 & \textbf{81.8} & 82.1 & 83.1 & 83.2
\\
\bottomrule
\end{tabularx}
\end{center}
% \vskip -0.1in
\end{table*}
\begin{table*}[b]
\caption{TPR at low FPR (FPR=5\%) results on the challenging MIMIR benchmark \protect\cite{duan2024membership} with non-member sets at $\le 20\%$ 7-gram overlap threshold with a suite of Pythia models \protect\cite{biderman2024emergent}. In each column, the best result across all methods is \textbf{bolded}. Despite not requiring a reference model like the Ref method does, our methods (AC, DerivAC, and NormAC) consistently achieve superior or comparable performance.
}
\label{tab:mimir_tpr_7_20}
\begin{center} \scriptsize
\setlength{\tabcolsep}{0.7pt}
\begin{tabularx}{\textwidth}{l *{20}{>{\centering\arraybackslash}X}@{}}
    \toprule
    \multirow{2}{*}{}  & \multicolumn{5}{c}{\textbf{Wikipedia}} & \multicolumn{5}{c}{\textbf{Github}} & \multicolumn{5}{c}{\textbf{PubMed Central}} \\
    \cmidrule(lr){2-6}  \cmidrule(lr){7-11} \cmidrule(lr){12-16}
    \textbf{Method}
    & 160M & 1.4B & 2.8B & 6.9B & 12B
    & 160M & 1.4B & 2.8B & 6.9B & 12B
    & 160M & 1.4B & 2.8B & 6.9B & 12B
    % & 160M & 1.4B & 2.8B & 6.9B & 12B
    \\
    \midrule
    
Loss \cite{yeom2018privacy}
& 19.1 & 23.2 & 22.8 & 24.2 & 24.4 & 44.0 & 57.1 & 56.7 & 60.4 & 62.3 & 28.1 & 31.0 & 32.2 & 29.7 & 33.0
\\
Ref \cite{carlini2021extracting}
& 11.2 & 21.2 & 19.8 & 20.4 & 21.2 & 36.2 & 26.9 & 23.9 & 25.4 & 20.5 & 19.1 & 5.5 & 4.1 & 5.1 & 4.5
\\
Lowercase \cite{carlini2021extracting}
& 20.7 & 21.4 & 21.5 & 22.8 & 23.9 & 63.4 & 67.5 & 67.5 & 66.0 & 67.9 & 21.4 & 25.3 & 18.9 & 20.0 & 17.1
\\
Zlib \cite{carlini2021extracting}
& 13.6 & 19.9 & 19.6 & 19.2 & 21.7 & 65.7 & 70.5 & 71.3 & 73.9 & 76.1 & 26.3 & 29.3 & 28.7 & 29.1 & 27.7
\\
Min-K\% \cite{shi2023detecting}
& 21.2 & 23.4 & 23.5 & 24.9 & 25.7 & 44.0 & 57.5 & 57.1 & 60.8 & 64.2 & 37.5 & 35.8 & 35.6 & 35.0 & 35.2
\\
Min-K\%++ \cite{zhang2024min}
& 16.8 & 19.4 & 22.6 & 23.4 & 25.9 & 43.3 & 51.9 & 56.0 & 53.7 & 54.1 & 22.4 & 21.0 & 20.8 & 23.6 & 21.8
\\
DC-PDD \cite{zhang2024pretraining}
& 21.3 & 26.9 & \textbf{27.9} & 27.3 & 28.8 & \textbf{73.1} & \textbf{76.9} & 78.7 & 79.1 & \textbf{79.9} & 45.2 & 42.0 & 40.5 & 38.9 & 37.5
\\
\hline
AC
& 21.1 & 27.6 & 27.0 & \textbf{28.7} & \textbf{30.2} & 72.8 & \textbf{76.9} & 78.7 & \textbf{79.5} & 79.1 & 45.8 & 43.6 & 41.1 & 39.7 & 39.3
\\
DerivAC
& \textbf{22.0} & \textbf{28.0} & 27.0 & 28.3 & 30.1 & \textbf{73.1} & \textbf{76.9} & \textbf{79.5} & \textbf{79.5} & 79.1 & \textbf{48.1} & \textbf{46.6} & \textbf{45.4} & 43.2 & \textbf{41.8}
\\
NormAC
& 20.2 & 24.7 & 25.3 & 26.1 & 27.7 & 67.2 & 75.4 & 75.4 & 77.6 & 78.4 & 47.0 & 44.8 & 44.6 & \textbf{47.9} & 40.3
\\

        \vspace{-.6em} \\
    
    \toprule
    \multirow{2}{*}{}  & \multicolumn{5}{c}{\textbf{ArXiv}} & \multicolumn{5}{c}{\textbf{DM Mathematics}} & \multicolumn{5}{c}{\textbf{Average}}\\
    \cmidrule(lr){2-6}  \cmidrule(lr){7-11} \cmidrule(lr){12-16}
    \textbf{Method} 
    & 160M & 1.4B & 2.8B & 6.9B & 12B
    & 160M & 1.4B & 2.8B & 6.9B & 12B
    & 160M & 1.4B & 2.8B & 6.9B & 12B
    % & 160M & 1.4B & 2.8B & 6.9B & 12B
    \\
    \midrule

Loss \cite{yeom2018privacy} 
& 29.4 & 32.2 & 34.2 & 36.6 & 37.0 & \textbf{82.0} & 62.9 & 59.6 & 64.0 & 61.8 & 40.5 & 41.3 & 41.1 & 43.0 & 43.7
\\
Ref \cite{carlini2021extracting} 
& 11.2 & 19.6 & 19.0 & 22.8 & 22.4 & 21.3 & 3.4 & 6.7 & 5.6 & 3.4 & 19.8 & 15.3 & 14.7 & 15.9 & 14.4
\\
Lowercase \cite{carlini2021extracting}
& 25.2 & 26.6 & 25.0 & 27.6 & 28.2 & 58.4 & 65.2 & 66.3 & 67.4 & 64.0 & 37.8 & 41.2 & 39.8 & 40.8 & 40.2
\\
Zlib \cite{carlini2021extracting}
& 27.8 & 30.4 & 35.0 & 32.0 & 33.6 & 25.8 & 16.9 & 16.9 & 16.9 & 16.9 & 31.8 & 33.4 & 34.3 & 34.2 & 35.2
\\
Min-K\% \cite{shi2023detecting}
& 32.2 & 33.2 & 36.8 & 37.4 & 39.4 & \textbf{82.0} & 69.7 & 70.8 & 70.8 & 71.9 & 43.4 & 43.9 & 44.8 & 45.8 & 47.3
\\
Min-K\%++ \cite{zhang2024min}
& 13.0 & 13.2 & 16.8 & 18.4 & 21.0 & 41.6 & 37.1 & 33.7 & 37.1 & 33.7 & 27.4 & 28.5 & 30.0 & 31.2 & 31.3
\\
DC-PDD \cite{zhang2024pretraining}
& 33.4 & 34.6 & 35.8 & 38.6 & 40.0 & 78.7 & 71.9 & 73.0 & 71.9 & 73.0 & 50.3 & 50.5 & 51.2 & 51.2 & 51.8
\\
\hline
AC
& 30.8 & 30.2 & 32.6 & 38.2 & 37.2 & 79.8 & 79.8 & 79.8 & 79.8 & 78.7 & 50.1 & 51.6 & 51.8 & 53.2 & 52.9
\\
DerivAC
& 37.4 & 39.2 & 39.0 & 41.8 & 43.4 & 77.5 & 77.5 & 78.7 & 78.7 & 78.7 & \textbf{51.6} & 53.6 & \textbf{53.9} & 54.3 & \textbf{54.6}
\\
NormAC
& \textbf{39.2} & \textbf{43.4} & \textbf{42.6} & \textbf{42.4} & \textbf{45.2} & 77.5 & \textbf{82.0} & \textbf{80.9} & \textbf{83.1} & \textbf{79.8} & 50.2 & \textbf{54.1} & 53.8 & \textbf{55.4} & 54.3
\\
    
    \bottomrule
\end{tabularx}
\end{center}
% \vskip -0.1in
\end{table*}
\begin{table*}[b]
\caption{AUROC results on the challenging MIMIR benchmark \protect\cite{duan2024membership} with non-member sets at $\le 80\%$ 13-gram overlap threshold with a suite of Pythia models \protect\cite{biderman2024emergent}.  In each column, the best result across all methods is \textbf{bolded}.
% Despite not requiring a reference model like the Ref method does, our methods (NormAR and AR) consistently achieves superior or comparable performance
}
\label{tab:mimir_auroc_13_80}
\begin{center} \scriptsize
\setlength{\tabcolsep}{0.7pt}
\begin{tabularx}{\textwidth}{l *{20}{>{\centering\arraybackslash}X}@{}}
\toprule
\multirow{2}{*}{}  & \multicolumn{5}{c}{\textbf{Wikipedia}} & \multicolumn{5}{c}{\textbf{Github}} & \multicolumn{5}{c}{\textbf{Pile CC}} & \multicolumn{5}{c}{\textbf{PubMed Central}} \\
\cmidrule(lr){2-6}  \cmidrule(lr){7-11} \cmidrule(lr){12-16} \cmidrule(lr){17-21}
\textbf{Method}
& 160M & 1.4B & 2.8B & 6.9B & 12B
& 160M & 1.4B & 2.8B & 6.9B & 12B
& 160M & 1.4B & 2.8B & 6.9B & 12B
& 160M & 1.4B & 2.8B & 6.9B & 12B
\\
\midrule

Loss \cite{yeom2018privacy}
& 50.2 & 51.3 & 51.7 & 52.7 & 53.3 & 65.7 & 69.8 & 71.2 & 72.9 & 73.9 & 49.5 & 50.0 & 50.1 & 50.7 & 51.0 & 50.0 & 49.8 & 49.9 & 50.6 & 51.1
\\
Ref \cite{carlini2021extracting}
& 50.8 & \textbf{55.2} & \textbf{57.9} & \textbf{61.1} & \textbf{63.0} & 64.7 & 67.1 & 65.5 & 65.2 & 63.9 & 49.1 & 52.2 & 53.5 & \textbf{54.8} & \textbf{56.2} & 51.2 & \textbf{53.1} & \textbf{53.6} & \textbf{55.8} & \textbf{57.9}
\\
Lowercase \cite{carlini2021extracting}
& 50.1 & 51.3 & 51.7 & 53.5 & 53.8 & 67.2 & 70.3 & 71.3 & 72.9 & 73.6 & 47.8 & 48.6 & 49.5 & 50.1 & 50.4 & 49.5 & 50.4 & 51.5 & 51.5 & 52.6
\\
Zlib \cite{carlini2021extracting}
& \textbf{50.9} & 52.0 & 52.4 & 53.5 & 54.2 & 67.4 & 71.0 & 72.3 & 73.8 & 74.7 & 49.5 & 50.1 & 50.3 & 50.8 & 51.1 & 50.1 & 50.0 & 50.1 & 50.6 & 51.0
\\
Min-K\% \cite{shi2023detecting}
& 50.2 & 51.3 & 51.8 & 53.5 & 54.3 & 65.7 & 69.9 & 71.4 & 73.1 & 74.1 & 50.4 & 51.0 & 50.8 & 51.4 & 51.7 & 50.9 & 50.3 & 50.4 & 51.2 & 52.2
\\
Min-K\%++ \cite{zhang2024min}
& 49.7 & 53.6 & 55.0 & 57.5 & 59.8 & 63.5 & 69.6 & 70.9 & 72.3 & 73.5 & 50.7 & 51.1 & 51.0 & 52.8 & 53.0 & 51.1 & 51.5 & 52.4 & 53.9 & 54.8
\\
DC-PDD \cite{zhang2024pretraining}
& \textbf{50.9} & 51.6 & 52.2 & 54.9 & 56.0 & 67.6 & 71.2 & 72.5 & 74.0 & 74.8 & 50.4 & 51.4 & 51.5 & 52.2 & 51.9 & 50.9 & 50.7 & 51.1 & 52.2 & 52.7
\\
\hline
AC
& 49.9 & 54.5 & 55.3 & 58.2 & 60.5 & 67.8 & \textbf{71.6} & \textbf{72.8} & \textbf{74.4} & \textbf{75.3} & 50.9 & 52.2 & 53.0 & 53.8 & 54.3 & 51.2 & 52.4 & 53.5 & 54.5 & 55.8
\\
DerivAC
& 50.2 & 54.5 & 55.3 & 58.2 & 60.4 & \textbf{68.3} & \textbf{71.6} & \textbf{72.8} & \textbf{74.4} & \textbf{75.3} & \textbf{51.4} & \textbf{52.7} & \textbf{53.6} & 53.8 & 54.2 & 51.3 & 52.4 & 53.5 & 54.6 & 55.8
\\
NormAC
& 50.2 & 54.7 & 55.5 & 58.0 & 60.0 & 67.5 & 70.7 & 71.8 & 73.0 & 74.0 & \textbf{51.4} & 52.0 & 53.2 & 53.2 & 53.4 & \textbf{51.6} & 52.6 & 53.3 & 54.5 & 55.6
\\

\vspace{-.6em} \\

\toprule
\multirow{2}{*}{}  & \multicolumn{5}{c}{\textbf{ArXiv}} & \multicolumn{5}{c}{\textbf{DM Mathematics}} & \multicolumn{5}{c}{\textbf{HackerNews}} & \multicolumn{5}{c}{\textbf{Average}}\\
\cmidrule(lr){2-6}  \cmidrule(lr){7-11} \cmidrule(lr){12-16} \cmidrule(lr){17-21}
\textbf{Method} 
& 160M & 1.4B & 2.8B & 6.9B & 12B
& 160M & 1.4B & 2.8B & 6.9B & 12B
& 160M & 1.4B & 2.8B & 6.9B & 12B
& 160M & 1.4B & 2.8B & 6.9B & 12B
\\
\midrule

Loss \cite{yeom2018privacy} 
& 50.7 & 51.5 & 51.9 & 52.7 & 53.3 & 48.9 & 48.5 & 48.4 & 48.4 & 48.5 & 49.5 & 50.5 & 51.3 & 51.9 & 52.8 & 52.1 & 53.1 & 53.5 & 54.3 & 54.8
\\
Ref \cite{carlini2021extracting} 
& 48.6 & 51.4 & 53.0 & 55.3 & 57.0 & 50.9 & 51.3 & 50.6 & 51.0 & 50.9 & 49.5 & \textbf{52.4} & \textbf{55.1} & \textbf{57.4} & \textbf{60.3} & 52.1 & 54.7 & 55.6 & 57.2 & 58.5
\\
Lowercase \cite{carlini2021extracting}
& 50.8 & 50.7 & 51.3 & 51.9 & 52.3 & 48.9 & 49.0 & 49.0 & 49.1 & 48.1 & 49.0 & 50.4 & 51.1 & 51.6 & 52.1 & 51.9 & 53.0 & 53.6 & 54.4 & 54.7
\\
Zlib \cite{carlini2021extracting}
& 49.9 & 50.8 & 51.3 & 52.1 & 52.5 & 48.0 & 48.2 & 48.0 & 48.0 & 48.0 & 49.8 & 50.3 & 50.8 & 51.1 & 51.7 & 52.2 & 53.2 & 53.6 & 54.3 & 54.7
\\
Min-K\% \cite{shi2023detecting}
& 50.7 & 51.6 & 52.5 & 53.5 & 54.4 & 49.3 & 49.7 & 49.5 & 49.6 & 49.7 & 51.3 & 51.3 & 52.4 & 53.4 & 54.8 & 52.6 & 53.6 & 54.1 & 55.1 & 55.9
\\
Min-K\%++ \cite{zhang2024min}
& 49.7 & 51.0 & 53.6 & 55.0 & \textbf{57.6} & 50.8 & 50.9 & 51.6 & 51.1 & 51.8 & \textbf{51.4} & 51.1 & 52.4 & 53.9 & 56.3 & 52.4 & 54.1 & 55.3 & 56.6 & 58.1
\\
DC-PDD \cite{zhang2024pretraining}
& \textbf{52.6} & 52.4 & 52.9 & 54.0 & 54.5 & 49.8 & 50.1 & 49.8 & 50.0 & 50.1 & 50.8 & 51.6 & 53.2 & 53.4 & 54.8 & 53.3 & 54.1 & 54.7 & 55.8 & 56.4
\\
\hline
AC
& 51.7 & 52.3 & 54.2 & \textbf{55.5} & 57.4 & \textbf{51.9} & 51.7 & 51.1 & 51.9 & \textbf{53.0} & 50.5 & 51.6 & 52.5 & 54.5 & 56.4 & \textbf{53.4} & 55.2 & 56.1 & 57.5 & \textbf{59.0}
\\
DerivAC
& 51.9 & 52.3 & 54.2 & \textbf{55.5} & 57.4 & 50.1 & 51.0 & 51.7 & 52.6 & 52.6 & 50.4 & 51.4 & 52.5 & 54.3 & 56.2 & \textbf{53.4} & 55.1 & \textbf{56.2} & \textbf{57.6} & 58.8
\\
NormAC
& 52.0 & \textbf{53.3} & \textbf{54.3} & \textbf{55.5} & 57.5 & 50.2 & \textbf{52.1} & \textbf{52.2} & \textbf{53.0} & 52.9 & 51.0 & 51.4 & 52.3 & 54.4 & 56.3 & \textbf{53.4} & \textbf{55.3} & 56.1 & 57.4 & 58.5
\\

\bottomrule
\end{tabularx}
\end{center}
% \vskip -0.1in
\end{table*}
\begin{table*}[b]
\caption{TPR at low FPR (FPR=5\%) results on the challenging MIMIR benchmark \protect\cite{duan2024membership} with non-member sets at $\le 80\%$ 13-gram overlap threshold with a suite of Pythia models \protect\cite{biderman2024emergent}.  In each column, the best result across all methods is \textbf{bolded}.
}
\label{tab:mimir_tpr_13_80}
\begin{center} \scriptsize
\setlength{\tabcolsep}{0.7pt}
\begin{tabularx}{\textwidth}{l *{20}{>{\centering\arraybackslash}X}@{}}
\toprule
\multirow{2}{*}{}  & \multicolumn{5}{c}{\textbf{Wikipedia}} & \multicolumn{5}{c}{\textbf{Github}} & \multicolumn{5}{c}{\textbf{Pile CC}} & \multicolumn{5}{c}{\textbf{PubMed Central}} \\
\cmidrule(lr){2-6}  \cmidrule(lr){7-11} \cmidrule(lr){12-16} \cmidrule(lr){17-21}
\textbf{Method}
& 160M & 1.4B & 2.8B & 6.9B & 12B
& 160M & 1.4B & 2.8B & 6.9B & 12B
& 160M & 1.4B & 2.8B & 6.9B & 12B
& 160M & 1.4B & 2.8B & 6.9B & 12B
\\
\midrule

Loss \cite{yeom2018privacy} 
& 4.2 & 4.7 & 4.6 & 5.0 & 5.0 & 21.9 & 32.2 & 33.8 & 37.8 & 40.6 & 3.1 & 4.8 & 4.8 & 4.9 & 5.5 & 3.3 & 4.5 & 4.4 & 4.9 & 5.3
\\
Ref \cite{carlini2021extracting} 
& 5.2 & 5.6 & 5.5 & 5.6 & 5.5 & 23.8 & 15.0 & 15.1 & 15.8 & 16.2 & 4.4 & 5.7 & 6.1 & 6.4 & 7.0 & 5.7 & 4.1 & 3.5 & 6.2 & 7.8
\\
Lowercase \cite{carlini2021extracting}
& 4.6 & 4.5 & 4.9 & 5.2 & 5.5 & 24.4 & 32.2 & 34.3 & 38.1 & 38.0 & 3.4 & 5.3 & 5.3 & 6.2 & 6.3 & 3.5 & 5.0 & 5.3 & 6.0 & 4.9
\\
Zlib \cite{carlini2021extracting}
& 4.8 & 5.6 & 5.9 & 6.3 & 6.9 & 24.3 & 32.8 & 36.0 & 38.7 & 41.2 & 3.6 & 5.0 & 5.3 & 5.6 & 6.7 & 3.4 & 3.7 & 3.6 & 4.2 & 4.4
\\
Min-K\% \cite{shi2023detecting}
& 5.8 & 5.7 & 6.0 & 6.9 & 7.2 & 22.0 & 32.2 & 34.2 & 37.8 & 40.6 & 4.0 & 5.1 & 5.0 & 5.3 & 5.8 & 5.0 & 4.6 & 4.7 & 5.7 & 5.7
\\
Min-K\%++ \cite{zhang2024min}
& 6.2 & 6.5 & \textbf{8.6} & 10.2 & 10.3 & 22.0 & 33.2 & 34.7 & 37.7 & 40.0 & \textbf{6.1} & 4.8 & 5.1 & 5.1 & 5.5 & \textbf{6.6} & 6.9 & 6.8 & 7.6 & 8.4
\\
DC-PDD \cite{zhang2024pretraining}
& \textbf{6.4} & 6.8 & 6.5 & 6.6 & 8.1 & 24.6 & 33.9 & \textbf{37.8} & 39.1 & \textbf{41.9} & 4.1 & 4.8 & 5.2 & 5.6 & 5.9 & 5.6 & 6.3 & 5.5 & 6.0 & 5.3
\\
\hline
AC
& 5.7 & 6.9 & 7.7 & 10.3 & 10.9 & 25.0 & 35.3 & 36.7 & \textbf{40.7} & \textbf{41.9} & 5.6 & 6.7 & \textbf{7.2} & 7.7 & 7.9 & 6.2 & \textbf{7.8} & \textbf{7.3} & 7.4 & \textbf{8.6}
\\
DerivAC
& 5.5 & 7.4 & 8.1 & \textbf{10.4} & \textbf{11.1} & \textbf{27.8} & 34.3 & 36.3 & 38.9 & 40.8 & 6.0 & 7.0 & \textbf{7.2} & 7.9 & \textbf{8.1} & 5.7 & 7.7 & \textbf{7.3} & 7.3 & \textbf{8.6}
\\
NormAC
& 5.7 & \textbf{8.3} & 8.1 & 10.3 & 10.5 & 24.7 & \textbf{36.1} & 36.7 & 38.6 & 40.1 & 5.9 & \textbf{7.2} & 6.7 & \textbf{8.0} & 7.9 & 6.2 & \textbf{7.8} & 6.7 & \textbf{8.6} & 8.4
\\

\vspace{-.6em} \\

\toprule
\multirow{2}{*}{}  & \multicolumn{5}{c}{\textbf{ArXiv}} & \multicolumn{5}{c}{\textbf{DM Mathematics}} & \multicolumn{5}{c}{\textbf{HackerNews}} & \multicolumn{5}{c}{\textbf{Average}}\\
\cmidrule(lr){2-6}  \cmidrule(lr){7-11} \cmidrule(lr){12-16} \cmidrule(lr){17-21}
\textbf{Method} 
& 160M & 1.4B & 2.8B & 6.9B & 12B
& 160M & 1.4B & 2.8B & 6.9B & 12B
& 160M & 1.4B & 2.8B & 6.9B & 12B
& 160M & 1.4B & 2.8B & 6.9B & 12B
\\
\midrule

Loss \cite{yeom2018privacy} 
& 4.3 & 4.6 & 4.7 & 5.1 & 5.6 & 3.7 & 4.2 & 4.1 & 4.2 & 3.9 & 4.8 & 4.8 & 6.0 & 5.6 & 6.3 & 6.5 & 8.5 & 8.9 & 9.6 & 10.3
\\
Ref \cite{carlini2021extracting} 
& 3.7 & 5.3 & 6.4 & 6.3 & 7.1 & 5.6 & 3.8 & 4.9 & 5.5 & 5.4 & 5.6 & 6.0 & \textbf{8.0} & 7.1 & \textbf{8.6} & 7.7 & 6.5 & 7.1 & 7.6 & 8.2
\\
Lowercase \cite{carlini2021extracting}
& 5.1 & 4.7 & 5.4 & 5.6 & 5.0 & 5.6 & 6.2 & 5.5 & 6.8 & 5.5 & 5.2 & 5.2 & 6.3 & 6.6 & 6.7 & 7.4 & 9.0 & 9.6 & 10.6 & 10.3
\\
Zlib \cite{carlini2021extracting}
& 2.8 & 4.2 & 4.2 & 4.5 & 4.2 & 4.1 & 5.0 & 4.7 & 4.2 & 4.3 & 5.3 & 5.6 & 5.6 & 5.5 & 5.7 & 6.9 & 8.8 & 9.3 & 9.9 & 10.5
\\
Min-K\% \cite{shi2023detecting}
& 5.3 & 4.7 & 4.7 & 5.4 & 5.7 & 4.9 & 4.5 & 4.7 & 4.7 & 5.4 & 6.4 & 5.8 & 6.0 & 6.0 & 6.7 & 7.6 & 8.9 & 9.3 & 10.3 & 11.0
\\
Min-K\%++ \cite{zhang2024min}
& \textbf{7.1} & 6.0 & 6.3 & 7.5 & 8.0 & 4.1 & 6.6 & 5.9 & 6.3 & 6.0 & 5.2 & 4.6 & 5.6 & 6.6 & 6.6 & 8.2 & 9.8 & 10.4 & 11.6 & 12.1
\\
DC-PDD \cite{zhang2024pretraining}
& 6.0 & 5.3 & 5.6 & 7.1 & 6.7 & 5.2 & 5.0 & 4.9 & 4.4 & 4.8 & 6.5 & 5.3 & 6.5 & 6.9 & 6.7 & 8.3 & 9.6 & 10.3 & 10.8 & 11.3
\\
\hline
AC
& 6.7 & 6.8 & \textbf{7.7} & \textbf{8.0} & \textbf{8.7} & 6.2 & 6.8 & 7.2 & 6.6 & 7.7 & 6.2 & 5.2 & 5.9 & 6.4 & 7.0 & 8.8 & 10.8 & 11.4 & 12.4 & \textbf{13.2}
\\
DerivAC
& \textbf{7.1} & 5.8 & 6.8 & 7.5 & 7.9 & 6.6 & 6.8 & 7.0 & 6.9 & 7.4 & \textbf{7.0} & 5.4 & 6.3 & 6.2 & 6.9 & \textbf{9.4} & 10.6 & 11.3 & 12.2 & 13.0
\\
NormAC
& 6.4 & \textbf{8.6} & 7.0 & 7.7 & 8.5 & \textbf{8.2} & \textbf{8.0} & \textbf{9.5} & \textbf{7.9} & \textbf{7.9} & 6.5 & \textbf{6.1} & 6.4 & \textbf{7.4} & 7.5 & 9.1 & \textbf{11.7} & \textbf{11.6} & \textbf{12.6} & 13.0
\\

\bottomrule
\end{tabularx}
\end{center}
% \vskip -0.1in
\end{table*}
\begin{table*}[b]
\caption{AUROC results on the challenging MIMIR benchmark \protect\cite{duan2024membership} with non-member sets at $\le 20\%$ 13-gram overlap threshold with a suite of Pythia models \protect\cite{biderman2024emergent}. In each column, the best result across all methods is \textbf{bolded}.
}
\label{tab:mimir_auroc_13_20}
\begin{center} \scriptsize
\setlength{\tabcolsep}{0.7pt}
\begin{tabularx}{\textwidth}{l *{20}{>{\centering\arraybackslash}X}@{}}
\toprule
\multirow{2}{*}{}  & \multicolumn{5}{c}{\textbf{Wikipedia}} & \multicolumn{5}{c}{\textbf{Github}} & \multicolumn{5}{c}{\textbf{Pile CC}} & \multicolumn{5}{c}{\textbf{PubMed Central}} \\
\cmidrule(lr){2-6}  \cmidrule(lr){7-11} \cmidrule(lr){12-16} \cmidrule(lr){17-21}
\textbf{Method}
& 160M & 1.4B & 2.8B & 6.9B & 12B
& 160M & 1.4B & 2.8B & 6.9B & 12B
& 160M & 1.4B & 2.8B & 6.9B & 12B
& 160M & 1.4B & 2.8B & 6.9B & 12B
\\
\midrule

Loss \cite{yeom2018privacy}
& 51.2 & 53.4 & 54.1 & 55.5 & 56.3 & 76.7 & 80.2 & 81.4 & 82.7 & 83.5 & 50.2 & 51.0 & 51.2 & 52.1 & 52.6 & 50.8 & 52.1 & 52.7 & 53.4 & 53.9
\\
Ref \cite{carlini2021extracting}
& 50.4 & \textbf{57.0} & \textbf{59.9} & \textbf{64.2} & \textbf{66.8} & 73.3 & 74.5 & 71.9 & 71.9 & 69.3 & 50.7 & \textbf{55.0} & \textbf{56.1} & \textbf{58.6} & \textbf{60.1} & 49.0 & 54.2 & \textbf{56.1} & \textbf{58.4} & \textbf{60.1}
\\
Lowercase \cite{carlini2021extracting}
& 50.9 & 53.9 & 54.4 & 56.4 & 56.9 & 79.8 & 81.6 & 82.4 & 83.8 & 84.4 & 50.6 & 51.5 & 52.3 & 53.3 & 54.5 & 52.2 & 54.1 & 54.6 & 55.3 & 55.4
\\
Zlib \cite{carlini2021extracting}
& 50.3 & 53.1 & 54.0 & 55.6 & 56.5 & 79.8 & 82.9 & 83.9 & 84.9 & 85.7 & 51.2 & 52.1 & 52.3 & 53.2 & 53.5 & 51.4 & 52.6 & 53.0 & 53.6 & 54.1
\\
Min-K\% \cite{shi2023detecting}
& 51.2 & 53.7 & 54.7 & 56.8 & 57.8 & 76.7 & 80.2 & 81.4 & 82.7 & 83.5 & 50.5 & 51.8 & 52.2 & 53.0 & 53.2 & 51.7 & 52.6 & 53.0 & 53.9 & 54.6
\\
Min-K\%++ \cite{zhang2024min}
& 51.8 & 55.4 & 57.0 & 59.7 & 61.8 & 73.4 & 78.3 & 79.7 & 80.8 & 81.9 & 50.2 & 53.4 & 53.1 & 54.5 & 54.8 & 49.6 & 52.7 & 53.4 & 55.1 & 55.8
\\
DC-PDD \cite{zhang2024pretraining}
& 52.1 & 54.3 & 55.1 & 57.3 & 58.7 & 81.5 & 84.0 & 84.9 & 86.0 & 86.6 & \textbf{52.3} & 53.2 & 53.7 & 53.9 & 54.4 & \textbf{52.4} & 53.4 & 53.9 & 54.2 & 55.3
\\
\hline
AC
& 51.9 & 55.6 & 56.4 & 59.4 & 61.6 & 81.2 & \textbf{84.3} & \textbf{85.1} & \textbf{86.1} & \textbf{86.8} & 51.9 & 54.8 & 55.0 & 56.0 & 55.7 & 51.6 & \textbf{54.3} & 55.3 & 56.3 & 57.7
\\
DerivAC
& 52.0 & 55.5 & 56.2 & 59.3 & 61.5 & \textbf{81.9} & 84.1 & 84.9 & 85.9 & 86.6 & 51.9 & 54.5 & 54.7 & 55.4 & 55.7 & 52.1 & \textbf{54.3} & 55.3 & 56.3 & 57.7
\\
NormAC
& \textbf{52.2} & 55.8 & 56.8 & 59.5 & 61.7 & 79.9 & 83.4 & 84.5 & 85.3 & 85.9 & 51.9 & 54.1 & 54.6 & 55.3 & 55.6 & \textbf{52.4} & 54.0 & 55.0 & 56.1 & 57.2
\\
\vspace{-.6em} \\
\toprule
\multirow{2}{*}{}  & \multicolumn{5}{c}{\textbf{ArXiv}} & \multicolumn{5}{c}{\textbf{DM Mathematics}} & \multicolumn{5}{c}{\textbf{HackerNews}} & \multicolumn{5}{c}{\textbf{Average}}\\
\cmidrule(lr){2-6}  \cmidrule(lr){7-11} \cmidrule(lr){12-16} \cmidrule(lr){17-21}
\textbf{Method} 
& 160M & 1.4B & 2.8B & 6.9B & 12B
& 160M & 1.4B & 2.8B & 6.9B & 12B
& 160M & 1.4B & 2.8B & 6.9B & 12B
& 160M & 1.4B & 2.8B & 6.9B & 12B
\\
\midrule
Loss \cite{yeom2018privacy} 
& 54.4 & 55.8 & 56.4 & 57.3 & 57.9 & \textbf{67.7} & \textbf{67.4} & \textbf{67.2} & \textbf{67.3} & \textbf{67.3} & 50.5 & 51.7 & 52.5 & 53.3 & 54.1 & 57.4 & 58.8 & 59.4 & 60.2 & 60.8
\\
Ref \cite{carlini2021extracting} 
& 50.7 & 54.8 & 56.2 & \textbf{58.6} & 59.9 & 58.9 & 47.2 & 45.2 & 45.0 & 44.4 & 49.6 & \textbf{52.4} & \textbf{55.1} & \textbf{57.8} & \textbf{60.8} & 54.7 & 56.4 & 57.2 & 59.2 & 60.2
\\
Lowercase \cite{carlini2021extracting}
& 53.7 & 54.2 & 55.2 & 55.8 & 56.7 & 56.5 & 56.6 & 58.6 & 58.0 & 56.7 & 49.3 & 50.6 & 51.5 & 52.1 & 52.3 & 56.1 & 57.5 & 58.4 & 59.2 & 59.6
\\
Zlib \cite{carlini2021extracting}
& 53.9 & 55.3 & 55.7 & 56.5 & 57.0 & 64.2 & 64.6 & 64.5 & 64.6 & 64.6 & 51.2 & 51.9 & 52.4 & 52.8 & 53.4 & 57.4 & 58.9 & 59.4 & 60.2 & 60.7
\\
Min-K\% \cite{shi2023detecting}
& 54.4 & 55.8 & 56.4 & 57.3 & 58.2 & \textbf{67.7} & \textbf{67.4} & \textbf{67.2} & \textbf{67.3} & \textbf{67.3} & 51.7 & 52.0 & 53.3 & 54.1 & 55.5 & 57.7 & 59.1 & 59.7 & 60.7 & 61.4
\\
Min-K\%++ \cite{zhang2024min}
& 52.4 & 54.3 & 56.3 & 57.6 & 59.9 & 57.5 & 58.6 & 59.6 & 59.2 & 59.2 & \textbf{51.9} & 51.4 & 52.9 & 54.4 & 56.8 & 55.3 & 57.7 & 58.9 & 60.2 & 61.5
\\
DC-PDD \cite{zhang2024pretraining}
& \textbf{55.8} & 56.4 & 56.4 & 57.5 & 58.4 & 62.6 & 63.1 & 62.7 & 62.9 & 62.8 & 50.7 & 51.7 & 53.3 & 54.1 & 55.2 & \textbf{58.2} & 59.4 & 60.0 & 60.8 & 61.6
\\
\hline
AC
& 54.7 & 56.3 & \textbf{57.2} & 58.5 & \textbf{60.1} & 62.6 & 64.2 & 64.0 & 64.0 & 63.9 & 51.0 & \textbf{52.4} & 52.9 & 54.8 & 56.7 & 57.8 & \textbf{60.3} & \textbf{60.8} & \textbf{62.2} & \textbf{63.2}
\\
DerivAC
& 55.1 & 56.3 & 57.0 & 58.4 & 59.9 & 63.2 & 64.2 & 64.2 & 64.1 & 63.6 & 50.5 & 52.0 & 53.0 & 54.7 & 56.4 & 58.1 & 60.1 & \textbf{60.8} & 62.0 & 63.1
\\
NormAC
& 55.5 & \textbf{56.5} & 56.8 & 57.3 & 59.5 & 61.0 & 64.5 & 64.6 & 64.3 & 64.0 & 51.1 & 52.1 & 52.8 & 54.7 & 56.6 & 57.7 & 60.1 & 60.7 & 61.8 & 62.9
\\
\bottomrule
\end{tabularx}
\end{center}
% \vskip -0.1in
\end{table*}
\begin{table*}[b]
\caption{TPR at low FPR (FPR=5\%) results on the challenging MIMIR benchmark \protect\cite{duan2024membership} with non-member sets at $\le 20\%$ 13-gram overlap threshold with a suite of Pythia models \protect\cite{biderman2024emergent}. In each column, the best result across all methods is \textbf{bolded}.
}
\label{tab:mimir_tpr_13_20}
\begin{center} \scriptsize
\setlength{\tabcolsep}{0.7pt}
\begin{tabularx}{\textwidth}{l *{20}{>{\centering\arraybackslash}X}@{}}
\toprule
\multirow{2}{*}{}  & \multicolumn{5}{c}{\textbf{Wikipedia}} & \multicolumn{5}{c}{\textbf{Github}} & \multicolumn{5}{c}{\textbf{Pile CC}} & \multicolumn{5}{c}{\textbf{PubMed Central}} \\
\cmidrule(lr){2-6}  \cmidrule(lr){7-11} \cmidrule(lr){12-16} \cmidrule(lr){17-21}
\textbf{Method}
& 160M & 1.4B & 2.8B & 6.9B & 12B
& 160M & 1.4B & 2.8B & 6.9B & 12B
& 160M & 1.4B & 2.8B & 6.9B & 12B
& 160M & 1.4B & 2.8B & 6.9B & 12B
\\
\midrule

Loss \cite{yeom2018privacy} 
& 8.0 & 10.1 & 11.7 & 11.7 & 12.3 & 38.4 & 44.7 & 50.0 & 53.4 & 54.5 & 3.4 & 7.5 & 7.3 & 8.3 & 8.4 & 4.4 & 4.8 & 5.6 & 5.9 & 6.2
\\
Ref \cite{carlini2021extracting} 
& 7.2 & 9.3 & 11.1 & 13.2 & 12.2 & 38.4 & 22.3 & 19.2 & 20.0 & 20.5 & 5.6 & 8.9 & \textbf{9.8} & \textbf{12.3} & \textbf{14.0} & 4.9 & 4.4 & 4.8 & \textbf{8.8} & 9.3
\\
Lowercase \cite{carlini2021extracting}
& 6.5 & 9.9 & 9.7 & 10.8 & 10.6 & 52.4 & 56.2 & 57.3 & 59.5 & 61.4 & 5.2 & 8.5 & 8.2 & 9.6 & 10.6 & 5.2 & 6.5 & 7.1 & 6.8 & 8.9
\\
Zlib \cite{carlini2021extracting}
& 7.0 & 8.5 & 9.5 & 10.0 & 11.1 & 53.4 & 58.1 & 59.9 & 62.2 & 63.0 & 6.1 & 7.3 & 8.1 & 9.2 & 9.4 & 4.0 & 4.6 & 5.1 & 5.5 & 6.1
\\
Min-K\% \cite{shi2023detecting}
& 8.6 & 10.1 & 11.8 & 12.7 & 13.2 & 38.4 & 44.9 & 50.1 & 53.4 & 54.5 & 4.6 & 7.6 & 7.4 & 8.3 & 8.6 & 4.5 & 6.0 & 5.8 & 6.5 & 6.9
\\
Min-K\%++ \cite{zhang2024min}
& 8.3 & 10.4 & 11.5 & 13.7 & 14.1 & 33.9 & 46.2 & 45.7 & 48.8 & 53.0 & 6.0 & \textbf{9.4} & 6.9 & 8.5 & 8.8 & 5.0 & 7.7 & 7.8 & 6.9 & 8.3
\\
DC-PDD \cite{zhang2024pretraining}
& 8.4 & 10.2 & 11.4 & 12.8 & 14.0 & 56.5 & 61.6 & 63.6 & \textbf{67.0} & 67.7 & 5.6 & 8.7 & 8.6 & 10.4 & 10.5 & 5.3 & 6.1 & 6.2 & 7.5 & 7.4
\\
\hline
AC
& \textbf{9.5} & \textbf{11.4} & 13.1 & \textbf{15.1} & 15.4 & 56.1 & 62.7 & 64.2 & 66.2 & 68.1 & 6.5 & 9.3 & 8.9 & 10.5 & 10.8 & 5.5 & \textbf{8.0} & 8.1 & 7.8 & 9.8
\\
DerivAC
& 9.0 & 10.2 & 13.1 & 14.7 & 15.2 & \textbf{60.0} & \textbf{63.5} & \textbf{65.3} & 66.6 & \textbf{68.6} & 6.4 & 9.3 & 8.8 & 10.5 & 10.4 & 5.9 & 7.6 & 7.7 & 7.6 & 9.8
\\
NormAC
& 8.5 & 10.3 & \textbf{13.4} & 13.6 & \textbf{15.5} & 53.2 & 61.2 & 63.5 & 65.5 & 67.7 & \textbf{7.0} & 9.0 & 8.6 & 10.3 & 10.8 & \textbf{6.3} & 7.7 & \textbf{8.5} & 8.4 & \textbf{11.1}
\\

\vspace{-.6em} \\

\toprule
\multirow{2}{*}{}  & \multicolumn{5}{c}{\textbf{ArXiv}} & \multicolumn{5}{c}{\textbf{DM Mathematics}} & \multicolumn{5}{c}{\textbf{HackerNews}} & \multicolumn{5}{c}{\textbf{Average}}\\
\cmidrule(lr){2-6}  \cmidrule(lr){7-11} \cmidrule(lr){12-16} \cmidrule(lr){17-21}
\textbf{Method} 
& 160M & 1.4B & 2.8B & 6.9B & 12B
& 160M & 1.4B & 2.8B & 6.9B & 12B
& 160M & 1.4B & 2.8B & 6.9B & 12B
& 160M & 1.4B & 2.8B & 6.9B & 12B
\\
\midrule

Loss \cite{yeom2018privacy} 
& 5.7 & 7.9 & 8.0 & 8.9 & 9.7 & 17.8 & 14.8 & 12.2 & 12.5 & 12.6 & 5.2 & 5.1 & 6.4 & 6.4 & 7.2 & 11.8 & 13.6 & 14.5 & 15.3 & 15.8
\\
Ref \cite{carlini2021extracting} 
& 5.9 & 8.4 & 8.6 & 9.9 & 10.4 & 7.1 & 2.2 & 3.0 & 3.0 & 2.7 & 5.9 & 6.0 & \textbf{8.2} & \textbf{7.9} & \textbf{9.4} & 10.7 & 8.8 & 9.2 & 10.7 & 11.2
\\
Lowercase \cite{carlini2021extracting}
& 7.9 & 7.7 & 7.5 & 9.5 & 9.3 & 10.4 & 9.5 & 9.4 & 15.3 & 10.8 & 5.0 & 5.1 & 6.5 & 6.8 & 6.8 & 7.8 & 7.4 & 7.8 & 10.5 & 9.0
\\
Zlib \cite{carlini2021extracting}
& 5.7 & 7.4 & 8.8 & 10.2 & 10.4 & 17.5 & 11.3 & 10.0 & 10.7 & 10.4 & 6.0 & 6.2 & 6.5 & 7.1 & 7.1 & 14.2 & 14.8 & 15.4 & 16.4 & 16.8
\\
Min-K\% \cite{shi2023detecting}
& 6.6 & 8.1 & 9.9 & 9.5 & 10.7 & 21.0 & 18.3 & 15.4 & 14.2 & 14.8 & 6.6 & 6.3 & 6.5 & 7.0 & 8.1 & 12.9 & 14.5 & 15.3 & 15.9 & 16.7
\\
Min-K\%++ \cite{zhang2024min}
& 6.9 & 7.9 & 8.4 & 10.5 & 12.1 & 7.1 & 11.1 & 12.2 & 12.5 & 12.1 & 6.0 & 5.2 & 5.4 & 6.7 & 6.6 & 10.5 & 14.0 & 14.0 & 15.4 & 16.4
\\
DC-PDD \cite{zhang2024pretraining}
& 8.1 & 9.3 & 9.8 & 10.6 & 11.7 & 15.6 & 15.6 & 14.5 & 15.4 & 16.1 & 6.6 & 5.8 & 7.1 & 6.7 & 6.9 & 15.2 & 16.8 & 17.3 & 18.6 & 19.2
\\
\hline
AC
& 8.2 & 9.7 & 10.5 & 11.0 & \textbf{12.5} & 18.0 & \textbf{20.7} & \textbf{19.7} & \textbf{21.6} & 18.7 & 6.6 & 5.2 & 5.8 & 6.6 & 7.7 & 15.8 & 18.1 & 18.6 & \textbf{19.8} & 20.4
\\
DerivAC
& \textbf{9.9} & \textbf{10.0} & \textbf{10.6} & \textbf{11.4} & 12.4 & \textbf{21.5} & \textbf{20.7} & 18.7 & 18.1 & 17.2 & \textbf{7.4} & 5.9 & 6.8 & 7.2 & 8.1 & \textbf{17.2} & \textbf{18.2} & \textbf{18.7} & 19.4 & 20.2
\\
NormAC
& 7.5 & 9.4 & 9.9 & 10.0 & 10.8 & 14.9 & 18.5 & 19.2 & 19.7 & \textbf{19.6} & 6.9 & \textbf{6.7} & 6.8 & 7.7 & 8.4 & 14.9 & 17.5 & 18.6 & 19.3 & \textbf{20.6}
\\

\bottomrule
\end{tabularx}
\end{center}
% \vskip -0.1in
\end{table*}

\end{document}